\documentclass[nohyperref]{article}

\usepackage{microtype}
\usepackage{graphicx}
\usepackage{subfigure}
\usepackage{booktabs} 

\usepackage{hyperref}

\usepackage[accepted]{icml2023}
\usepackage{algorithmic}

\usepackage{amsmath}
\usepackage{amssymb}
\usepackage{mathtools}
\usepackage{amsthm}

\usepackage[capitalize,noabbrev]{cleveref}

\theoremstyle{plain}
\newtheorem{theorem}{Theorem}[section]

\theoremstyle{definition}

\theoremstyle{remark}

\usepackage{booktabs}
\usepackage{graphicx}
\usepackage{bm}
\usepackage{amsmath}
\usepackage{amsfonts}
\usepackage{adjustbox}
\usepackage{bbm}
\usepackage{multirow}
\usepackage{amsthm}
\usepackage{xcolor}
\usepackage[misc]{ifsym}
\usepackage{bbding}

\usepackage[textsize=tiny]{todonotes}
\newcommand\bmt[1]{\bm{\tilde{#1}}}

\icmltitlerunning{Bit Allocation using Optimization}

\begin{document}

\twocolumn[
\icmltitle{Bit Allocation using Optimization}



\icmlsetsymbol{equal}{*}

\begin{icmlauthorlist}
\icmlauthor{Tongda Xu}{equal,atu}
\icmlauthor{Han Gao}{equal,str,uestc}
\icmlauthor{Chenjian Gao}{str,buaa}
\icmlauthor{Yuanyuan Wang}{str}
\icmlauthor{Dailan He}{str}
\icmlauthor{Jinyong Pi}{str}
\icmlauthor{Jixiang Luo}{str}
\icmlauthor{Ziyu Zhu}{ctu}
\icmlauthor{Mao Ye}{uestc}
\icmlauthor{Hongwei Qin}{str}
\icmlauthor{Yan Wang \Envelope}{atu}
\icmlauthor{Jingjing Liu}{atu,mtu}
\icmlauthor{Ya-Qin Zhang}{atu,ctu,mtu}
\end{icmlauthorlist}

\icmlaffiliation{atu}{Institute for AI Industry Research (AIR), Tsinghua University}
\icmlaffiliation{str}{SenseTime Research}
\icmlaffiliation{uestc}{University of Electronic Science and Technology of China}
\icmlaffiliation{buaa}{Beihang University}
\icmlaffiliation{mtu}{School of Vehicle and Mobility, Tsinghua University}
\icmlaffiliation{ctu}{Department of Computer Science and Technology, Tsinghua University}

\icmlcorrespondingauthor{Yan Wang}{wangyan@air.tsinghua.edu.cn}

\icmlkeywords{Machine Learning, ICML}

\vskip 0.3in
]



\printAffiliationsAndNotice{\icmlEqualContribution} 

\begin{abstract}
In this paper, we consider the problem of bit allocation in Neural Video Compression (NVC). First, we reveal a fundamental relationship between bit allocation in NVC and Semi-Amortized Variational Inference (SAVI). Specifically, we show that SAVI with GoP (Group-of-Picture)-level likelihood is equivalent to pixel-level bit allocation with precise rate \& quality dependency model. Based on this equivalence, we establish a new paradigm of bit allocation using SAVI. Different from previous bit allocation methods, our approach requires no empirical model and is thus optimal. Moreover, as the original SAVI using gradient ascent only applies to single-level latent, we extend the SAVI to multi-level such as NVC by recursively applying back-propagating through gradient ascent. Finally, we propose a tractable approximation for practical implementation. Our method can be applied to scenarios where performance outweights encoding speed, and serves as an empirical bound on the R-D performance of bit allocation. Experimental results show that current state-of-the-art bit allocation algorithms still have a room of $\approx 0.5$ dB PSNR to improve compared with ours. Code is available at {\small \url{https://github.com/tongdaxu/Bit-Allocation-Using-Optimization}}.
\end{abstract}

\section{Introduction}

Neural Video Compression (NVC) has been an active research area. Recently, state-of-the-art (SOTA) NVC approaches \citep{hu2022coarse,10.1145/3503161.3547845} have achieved comparable performance with advanced traditional video coding standards such as H.266 \citep{bross2021developments}. The majority of works in NVC focus on improving motion representation \citep{lu2019dvc,lu2020end,agustsson2020scale} and better temporal context \citep{Djelouah_2019_ICCV,lin2020m,yang2020hierarchical,yilmaz2021end,li2021deep}. However, the bit allocation of NVC is relatively under-explored \citep{li2022rate}.

The target of video codec is to minimize R-D (Rate-Distortion) cost $R+\lambda D$, where $R$ is bitrate, $D$ is distortion and $\lambda$ is the Lagrangian multiplier controlling R-D trade-off.
Due to the frame reference structure of video coding, using the same $\lambda$ for all frames/regions is suboptimal. Bit allocation is the task of solving $\lambda$ for different frames/regions. For traditional codecs, the accurate bit allocation has been considered intractable. And people solve $\lambda$ approximately via empirical rate \& quality dependency model \citep{li2014lambda,li2016lambda} (See details in Sec.~\ref{sec:lam})).

The pioneer of bit allocation for NVC \citep{rippel2019learned,li2022rate} adopts the empirical rate dependency from \citet{li2014lambda} and proposes a quality dependency model based on the frame reference relationship. More recently, \citet{10.1145/3503161.3547845} propose a feed-forward bit allocation approach with empirical dependency modeled implicitly by neural network. However, the performance of those approaches heavily depends on the accuracy of empirical model. On the other hand, we show that an earlier work, Online Encoder Update (OEU) \citep{lu2020content}, is in fact also a frame-level bit allocation for NVC (See Appendix.~\ref{app:oeu}).
Other works adopt simplistic heuristics such as fixed $\lambda$ schedule to achieve very coarse bit allocation \citep{cetin2022flexible,hu2022coarse,li2023neural}.

In this paper, we first examine the relationship of bit allocation in NVC and Semi-Amortized Variational Inference (SAVI) \citep{kim2018semi,marino2018iterative}. We prove that SAVI using GoP-level likelihood is equivalent to pixel-level bit allocation using precise rate \& quality dependency model. Based on this relationship, we propose a new paradigm of bit allocation using SAVI. Different from previous bit allocation methods, this approach achieves pixel-level control and requires no empirical model. And thus, it is optimal assuming gradient ascent can achieve global maxima. Moreover, as the original SAVI using gradient ascent only applies to single-level latent variable, we extend SAVI to latent with dependency by recursively applying back-propagating through gradient ascent. Furthermore, we provide a tractable approximation to this algorithm for practical implementation. Despite our approach increases encoding complexity, it does not affect decoding complexity. Therefore, it is applicable to scenarios where R-D performance is more important than encoding time. And it also serves as an empirical bound on the R-D performance of other bit allocation methods. Experimental results show that current bit allocation algorithms still have a room of $\approx 0.5$ dB PSNR to improve, compared with our results. Our bit allocation method is compatible with any NVC method with a differentiable decoder. And it can be even directly adopted on existing pre-trained models.

To wrap up, our contributions are as follows:
\begin{itemize}
    \item We prove the equivalence of SAVI on NVC with GoP-level likelihood and pixel-level bit allocation using the precise rate \& quality dependency model.
    \item We establish a new paradigm of bit allocation algorithm using gradient based optimization. Unlike previous methods, it requires no empirical model and is thus optimal.
    \item We extend the original SAVI to latent with general dependency by recursively applying back-propagating through gradient ascent. And we further provide a tractable approximation so it scales to practical problems such as NVC.
    \item Empirical results verify that the current bit allocation algorithms still have a room of $\approx 0.5$ dB PSNR to improve, compared with our optimal results.
\end{itemize}

\section{Preliminaries}
\subsection{Neural Video Compression}
The majority of NVC follow a mixture of latent variable model and temporal autoregressive model \citep{yang2020hierarchical}. To encode $i^{th}$ frame $\bm{x}_i \in \mathbb{R}^{HW}$ with $HW$ pixels inside a GoP $\bm{x}_{1:N}$ with N frames, we first transform the $i^{th}$ frame in context of previous frames to obtain latent parameter $\bm{y}_i=f_{\phi}(\bm{x}_i,\lfloor\bm{y}_{<i}\rceil)$, where $f_{\phi}(\cdot)$ is the encoder (inference model) parameterized by $\phi$, $\lfloor\cdot\rceil$ is the rounding operator, and $\lfloor\bm{y}_{<i}\rceil$ is the quantized latent of previous frames. Then, we obtain $\lfloor\bm{y}_{i}\rceil$ by quantizing the latent $\bm{y}_i$. Next, an entropy model $p_{\theta}(\lfloor\bm{y}_{i}\rceil|\lfloor\bm{y}_{<i}\rceil)$ parameterized by $\theta$ is used to evaluate the probability mass function (pmf, prior) $P_{\theta}(\lfloor\bm{y}_{i}\rceil|\lfloor\bm{y}_{<i}\rceil)=F_{\theta}(\lfloor\bm{y}_{i}\rceil+0.5|\lfloor\bm{y}_{<i}\rceil)-F_{\theta}(\lfloor\bm{y}_{i}\rceil-0.5|\lfloor\bm{y}_{<i}\rceil)$, where $F_{\theta}$ is the cdf of $p_{\theta}$. With the pmf, we encode $\lfloor\bm{y}_{i}\rceil$ with bitrate $R_i=-\log P_{\theta}(\lfloor\bm{y}_{i}\rceil|\lfloor\bm{y}_{<i}\rceil)$. During decoding process, we obtain the reconstruction $\hat{\bm{x}}_i=g_{\theta}(\lfloor\bm{y}_{\le i}\rceil)$, where $g_{\theta}(\cdot)$ is the decoder (generative model) parameterized by $\theta$. Finally, we compute distortion $D_i=d(\bm{x}_i,\hat{\bm{x}}_i)$, which can be interpreted as the data likelihood $\log p_{\theta}(\bm{x}_i|\lfloor\bm{y}_{\le i}\rceil)$ so long as we treat $D_i$ as energy of Gibbs distribution \citep{minnen2018joint}. For example, when $d(\cdot,\cdot)$ is mean square error (MSE), we can interpret $d(\bm{x},\bm{\hat{x}})=-\log p_{\theta}(\bm{x}|\lfloor \bm{y}\rceil)+C$, where $p_{\theta}(\bm{x}|\lfloor \bm{y}\rceil)=\mathcal{N}(\bm{\hat{x}},1/2\lambda_i I)$. The optimization target is the frame-level R-D cost (Eq.~\ref{eq:rd0}) with Lagrangian multiplier $\lambda$ controlling R-D trade-off. In this paper, we consider pixel-level distortion and Lagrangian multiplier, and we denote $\bm{D}_i\in \mathbb{R}^{HW}$ as the pixel-level distortion, $\bm{\lambda}_i\in \mathbb{R}^{HW}$ as the pixel-level Lagrangian multiplier, and $\bm{\lambda}_i^T\bm{D}_i$ as the weighted distortion. The above procedure can be described by:
\begin{gather}
    \bm{y}_i = f_{\phi}(\bm{x}_i, \bmt{y}_{<i})\textrm{, where } \bmt{y}_i=\lfloor\bm{y}_i\rceil, \label{eq:enc}\\
    R_i=-\log P_{\theta}(\bmt{y}_i|\bmt{y}_{<i})\textrm{, }\bm{D}_i=d(\bm{x}_i,g_{\theta}(\bmt{y}_{\le i})), \label{eq:dec} \\
    \phi^{*},\theta^{*} \leftarrow \arg \min_{\phi,\theta} R_i + \bm{\lambda}^T_i \bm{D}_i. \label{eq:rd0}
\end{gather}
As the rounding operation $\lfloor\cdot\rceil$ is not differentiable, \citet{balle2016end} propose to relax it by additive uniform noise (AUN), and replace $\lfloor\bm{y}_i\rceil$ with $\tilde{\bm{y}}_i=\bm{y}_i+\mathcal{U}(-1/2,1/2)$ during training. Under such formulation, we can also treat the above encoding-decoding process as a Variational Autoencoder \citep{kingma2013auto} on graphic model $\tilde{\bm{y}}_{\le i} \rightarrow \bm{x}_i$ with variational posterior: 
\begin{gather}
     q_{\phi}(\tilde{\bm{y}}_{1:N}|\bm{x}_{1:N}) = \prod_{i=1}^{N} q_{\phi}(\tilde{\bm{y}}_i|\bm{x}_i,\tilde{\bm{y}}_{<i}), \notag\\
    q_{\phi}(\tilde{\bm{y}}_i|\bm{x}_i,\tilde{\bm{y}}_{<i}) = \mathcal{U}(\bm{y}_i-1/2,\bm{y}_i+1/2)\label{eq:pos}.
\end{gather}
And then minimizing the uniform noise relaxed R-D cost (Eq.~\ref{eq:rd0}) is equivalent to maximizing the evident lowerbound (ELBO) (Eq.~\ref{eq:elbo}) by Stochastic Gradient Variational Bayes-A (SGVB-A) \citep{kingma2013auto}:
\begin{align}
    \mathcal{L}_i =& \mathbb{E}_{q_{\phi}(\tilde{\bm{y}}_i|\bm{x}_i,\tilde{\bm{y}}_{<i})}[\underbrace{\log P_{\theta}(\tilde{\bm{y}}_{i}|\tilde{\bm{y}}_{<i})}_{-R_i} +\underbrace{\log p_{\theta}(\bm{x}_i|\tilde{\bm{y}}_{\le i})}_{-\lambda D_i} \notag\\&-\underbrace{\log q_{\phi}(\tilde{\bm{y}}_i|\bm{x}_i,\tilde{\bm{y}}_{<i})}_{\textrm{bits-back bitrate: } 0}].
    \label{eq:elbo}
\end{align}

\subsection{$\lambda$-Domain Bit Allocation}
\label{sec:lam}
In this section, we briefly review the main results of $\lambda$-domain bit allocation \citep{li2016lambda} (See detailed derivation in Appendix.~\ref{app:lamb}). Consider the task of compressing $\bm{x}_{1:N}$ with a overall R-D trade-off parameter $\lambda_0$. Then the target of pixel-level bit allocation is to minimize GoP-level R-D cost by adjusting the pixel-level R-D trade-off parameter $\bm{\lambda}_i$:
\begin{gather}
    \bm{\lambda}_1^*,\cdots,\bm{\lambda}_N^* \leftarrow \arg \min_{\bm{\lambda}_1,\cdots,\bm{\lambda}_N} \sum_{i=1}^N R_i^* + \bm{\lambda}^T_0 \sum_{i=1}^N \bm{D}_i^*,\notag \\
    \textrm{where } R_i^*, \bm{D}_i^* \leftarrow \min R_i + \bm{\lambda}_i^T \bm{D}_i,\label{eq:rd1}
\end{gather}

where $\bm{\lambda}_0=\lambda_{0}I$ and $I$ is identity matrix. And following \citet{li2016lambda} and \citet{tishby2000information}, we have the optimal conditions for $\bm{\lambda}_i$ as:
\begin{gather}
    \sum_{j=i}^N\frac{d R_j}{d \bm{\lambda}_i}+\bm{\lambda}_0^T \sum_{j=i}^N\frac{d \bm{D}_j}{d \bm{\lambda}_i} = 0, \label{eq:rdgrad} \\
    \bm{\lambda}^T_i+\frac{d R_i}{d \bm{D}_i} =0. \label{eq:rdgradl}
\end{gather}
The conventional $\lambda$-domain bit allocation \citep{li2016lambda} introduces empirical models to solve $\bm{\lambda}_i$. Specifically, it approximates the rate \& quality dependency as:
\begin{align}
    \frac{d R_{j\neq i}}{d R_i} \approx 0,
    \sum_{j=i}^N \frac{d\bm{D}_j}{d\bm{D}_i} \approx \omega_iI,\label{eq:dep}
\end{align}
where $\omega_i$ is the model parameter. By taking Eq.~\ref{eq:dep} into Eq.~\ref{eq:rdgrad} (See details in Appendix.~\ref{app:lamb}), we have:
\begin{align}
    \frac{d R_i}{d \bm{\lambda}_i}+\omega_i \bm{\lambda}_0^T \frac{d \bm{D}_i}{d \bm{\lambda}_i} \approx 0. \label{eq:lamcon}
\end{align}
By taking Eq.~\ref{eq:rdgradl} into Eq.~\ref{eq:lamcon} (See details in Appendix.~\ref{app:lamb}), we can solve the pixel-level R-D trade-off parameter as:
\begin{align}
    \bm{\lambda}_i = \omega_i \bm{\lambda}_0 \approx \bm{\lambda}_i^*.\label{eq:lamlam}
\end{align}
As $\bm{\lambda}_i$ is only an approximation to the optimal solution $\bm{\lambda}_i^*$ in Eq.~\ref{eq:rd1}, the performance of $\lambda$-domain bit allocation heavily relies on the correctness of the empirical rate \& quality dependency model in Eq.~\ref{eq:dep}. 
\section{Bit Allocation using Optimization}
\subsection{Bit Allocation and SAVI}
\label{sec:ba}
Consider applying SAVI \citep{kim2018semi} to NVC with GoP-level likelihood as target. This means that we initialize the variational posterior parameters $\bm{y}_{1:N}$ from amortized encoder $f_{\phi}(\bm{y}_{1:N}|\bm{x})$, and optimize them directly to maximize GoP-level ELBO or minus GoP-level R-D cost with $\bm{\lambda}_0$ as trade-off parameter:
\begin{gather}
    \bm{y}^*_{1:N} \leftarrow \arg \max_{\bm{y}_{1:N}} \mathcal{L},\notag \\
    \textrm{where }\mathcal{L} = \overset{N}{\underset{i=1}{\sum}}\mathcal{L}_{i} = \overset{N}{\underset{i=1}{\sum}}-(R_i+\bm{\lambda}^T_0 \bm{D}_i). 
    \label{eq:cee6}
\end{gather}
It is not surprising that this optimization can improve the R-D performance of NVC, as previous works in density estimation \citep{kim2018semi,marino2018iterative} and image compression \citep{yang2020improving,gaoflexible} have shown that this approach can reduce the amortization gap of VAE \citep{cremer2018inference}, and thus reduce R-D cost.

 However, it is interesting to understand its relationship to bit allocation. First, let's construct a pixel-level bit allocation map $\bm{\lambda}^{'}_{i}$ that satisfies:
\begin{align}
    \frac{d -(R_i+\bm{\lambda}^{'T}_{i}\bm{D}_i)}{d \bm{y}_i} \propto \frac{d \mathcal{L}}{d \bm{y}_i}.\label{eq:cee7}
\end{align}
We call $\bm{\lambda}^{'}_{i}$ the equivalent bit allocation map, as minimizing the single frame R-D cost $R_i+\bm{\lambda}^{'T}_{i}\bm{D}_i$ is equivalent to maximizing the GoP-level likelihood $\mathcal{L}$. In other words, $\bm{\lambda}_i^{'}$ is the bit allocation that is equivalent to SAVI using GoP-level likelihood. Moreover, $\bm{\lambda}_i^{'}$ can be solved explicitly, and it is the solution to optimal bit allocation problem of Eq.~\ref{eq:rd1}. Formally, we have the following results:
\begin{theorem}
\label{th:eqls}
The SAVI using GoP-level likelihood is equivalent to bit allocation with:
\begin{align}
    \bm{\lambda}^{'}_i=&(I+\sum_{j=i+1}^{N}\frac{d \bm{D}_j}{d \bm{D}_{i}})^T\bm{\lambda}_0/(1+\sum_{j=i+1}^{N}\frac{d R_j}{d R_i}), \label{eq:lstar}
\end{align}
where $d \bm{D}_j/d \bm{D}_{i},d R_j/d R_{i}$ can be computed numerically through the gradient of NVC model.
(See proof in Appendix.~\ref{app:proof})
\end{theorem}
\begin{theorem}
\label{th:eqrd}
The equivalent bit allocation map $\bm{\lambda}^{'}_{i}$ is the solution to the optimal bit allocation problem in Eq.~\ref{eq:rd1}. In other words, we have:
\begin{align}
    \bm{\lambda}_i^{'} = \bm{\lambda}_i^*.\label{eq:leq}
\end{align}
(See proof in Appendix.~\ref{app:proof})
\end{theorem}

Theorem.~\ref{th:eqrd} shows that SAVI with GoP-level likelihood is equivalent to pixel-level accurate bit allocation. Compared with previous bit allocation methods, this SAVI-based bit allocation does not require empirical model, and thus its solution $\bm
{\lambda}_i^{'}$ is accurate instead of approximated. Furthermore, it generalizes the $\lambda$-domain approach's solution $\bm{\lambda}_i$ in Eq.~\ref{eq:lamlam} (See details in Appendix.~\ref{app:lamb}).

Given external $\bm{\lambda}_i^{'}$, we can achieve bit allocation by optimizing the frame-level R-D cost. However, it is important to note that $\bm{\lambda}_i^{'}$ is intractable, as it requires matrix multiplication between quality dependency model $d \bm{D}_j/d \bm{D}_i$). A single precision quality dependency model of a $1280\times720$ video costs $3,397$ GB to store. 

On the other hand, the result in Theorem.~\ref{th:eqls} is also correct for traditional codec. However, as $\bm{\lambda}_i^{'}$ is intractable and the traditional codec is not differentiable, we can not use SAVI to achieve bit allocation for traditional codec.

\subsection{A Na\"ive Implementation}
The most intuitive way to achieve the SAVI/bit allocation described above is directly updating all latent with gradient ascent:
\begin{gather}
    \bm{y}^{0}_{i} \leftarrow f_{\phi}(\bm{x},\bm{y}^{0}_{<i}) \textrm{, }
    \bm{y}^{k+1}_{i} \leftarrow \bm{y}^{k}_{i} + \alpha \frac{d \mathcal{L}(\bm{y}^{k}_{1:N})}{d \bm{y}^{k}_{i}},\notag\\
    \textrm{where } \frac{d \mathcal{L}(\bm{y}^{k}_{1:N})}{d \bm{y}^{k}_{i}} = \sum_j^{N}\frac{\partial \mathcal{L}_j(\bm{y}^{k}_{j:N})}{\partial \bm{y}^{k}_{i}}.
    \label{eq:gradnaive}
\end{gather}
More specifically, we first initialize the variational posterior parameter $\bm{y}_i^0$ by fully amortized variational inference (FAVI) encoder $f_{\phi}(\bm{x},\bm{y}_{<i}^0)$. Then we iteratively update the $k^{th}$ step latent $\bm{y}_i^{k}$ with gradient ascent and learning rate $\alpha$ for $K$ steps. During this process, all the posterior parameter $\bm{y}^k_{1:N}$ are synchronously updated. This procedure is described in Alg.~\ref{alg:naive}.
\begin{algorithm}[H]
    \caption{Original SAVI on DAG}
    \label{alg:naive}
\begin{algorithmic}
    \STATE {\bfseries procedure} solve-original($\bm{x}$)
    \STATE initialize $\bm{y}_1^0,...,\bm{y}_N^0\leftarrow f_{\phi}(\bm{x})$ from FAVI.
    \FOR{$k=0$ {\bfseries to} $K-1$}
    \FOR{$i=1$ {\bfseries to} $N$}
    \STATE $\bm{y}_i^{k+1}\leftarrow\bm{y}_i^k+\alpha\frac{\partial\mathcal{L}(\bm{y}_1^k,...,\bm{y}_N^k)}{\partial\bm{y}_i^k}$
    \ENDFOR
    \ENDFOR
    \STATE {\bfseries return} $\bm{y}_1^K,...,\bm{y}_N^K$
\end{algorithmic}
\end{algorithm}
In fact, Alg.~\ref{alg:naive} is exactly the procedure of original SAVI \citep{kim2018semi} without amortized parameter update. This algorithm is designed for single-level variational posterior (the factorization used in mean-field \citep{blei2017variational}), which means $q_{\phi}(\tilde{\bm{y}}_{1:N}|\bm{x}_{1:N}) = \prod_{i=1}^{N} q_{\phi}(\tilde{\bm{y}}_i|\bm{x}_{1:N})$. However, the variational posterior of NVC has an autoregressive form (See Eq.~\ref{eq:pos}), which means that later frame's posterior parameter $\bm{y}_{>i}$ depends on current frame's parameter $\bm{y}_i$.

\subsection{The Problem with the Na\"ive Implementation}
\label{sec:problem}
For latent with dependency such as NVC, Alg.~\ref{alg:naive} becomes problematic. The intuition is, when computing the gradient for current frame's posterior parameter $\bm{y}_i$, we need to consider $\bm{y}_i$'s impact on the later frame $\bm{y}_{>i}$. And abusing SAVI on non-factorized latent causes gradient error in two aspects: (1). The total derivative $d \mathcal{L}/d\bm{y}_i$ is incomplete. (2). The total derivative $d \mathcal{L}/d\bm{y}_i$ and partial derivative $\partial\mathcal{L}_j/\partial\bm{y}_i$ are evaluated at wrong value. 
\begin{figure}[thb]
\centering
 \includegraphics[width=0.9\linewidth]{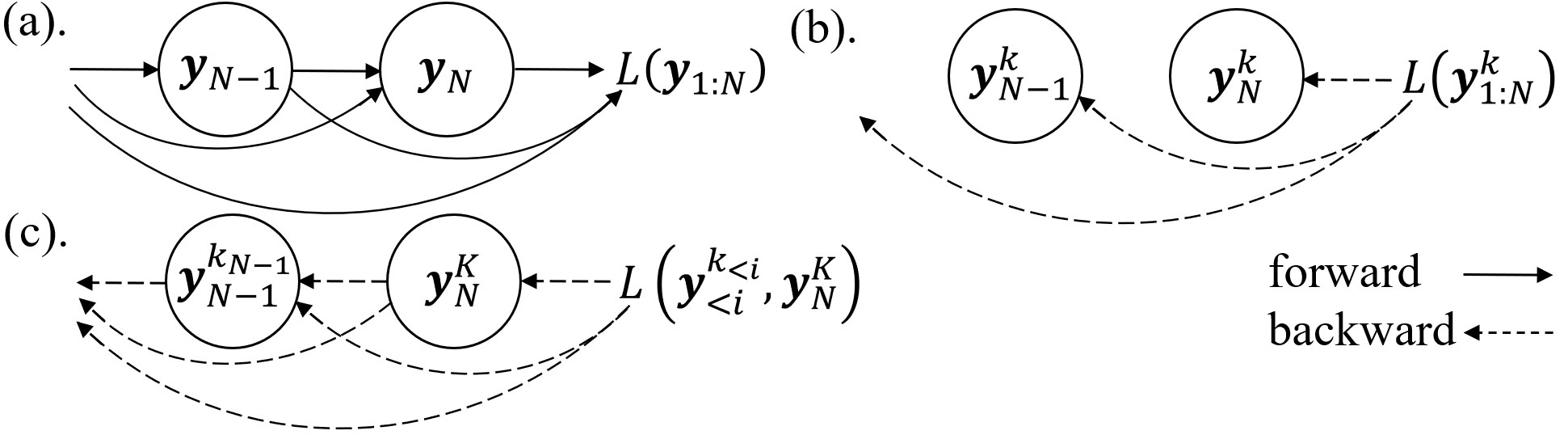}
 \caption{(a). The forward pass of NVC. (b). The backward pass of Na\"ive implementation (Alg.~\ref{alg:naive}). (c). The backward pass of advanced implementation (Alg.~\ref{alg:advancedag}).}
 \label{fig:cg}
\end{figure}

\textbf{Incomplete Total Derivative Evaluation} According to the latent's autogressive dependency in Eq.~\ref{eq:enc} and target $\mathcal{L}$ in Eq.~\ref{eq:cee6}, we draw the computational graph to describe the latent dependency as Fig.~\ref{fig:cg}.(a) and expand the total derivative $d\mathcal{L}/d\bm{y}_i$:
\begin{align}
\frac{d \mathcal{L}(\bm{y}_{1:N})}{d \bm{y}_i}=&\sum_{j=i}^{N}(\underbrace{\sum_{l=i+1}^{j}\frac{\partial \bm{y}_{l}}{\partial\bm{y}_i}\frac{d \mathcal{L}_j(\bm{y}_{1:j})}{d \bm{y}_l}}_{\textrm{ignored by na\"ive implementation}}+\frac{\partial \mathcal{L}_j(\bm{y}_{1:j})}{\partial \bm{y}_i}).\label{eq:incyg}
\end{align}
As shown in Eq.~\ref{eq:gradnaive} and Alg.~\ref{alg:naive}, The na\"ive implementation treats the total derivative $d\mathcal{L}/d\bm{y}_i$ as the sum of the frame level partial derivative $\partial \mathcal{L}_j/\partial\bm{y}_i$, which is the direct contribution of frame $i^{th}$ latent $\bm{y}_i$ to $j^{th}$ frame's R-D cost $\mathcal{L}_j$ (as marked in Eq.~\ref{eq:incyg}). This incomplete evaluation of gradient signal brings sub-optimality.

\textbf{Incorrect Value to Evaluate Gradient} Besides the incomplete gradient issue, Alg.~\ref{alg:naive} simultaneously updates all the posterior parameter $\bm{y}_{1:N}$ with gradient evaluated at the same step $\bm{y}_{1:N}^k$. However, to evaluate the gradient of $\bm{y}_i$, all its descendant latent $\bm{y}_{>i}$ should already complete all $K$ steps of gradient ascent. Moreover, once $\bm{y}_i$ is updated, all its descendant latent $\bm{y}_{> i}$ should be re-initialized by FAVI. Specifically, the correct value to evaluate the gradient is:
\begin{gather}
     \bm{y}^{k_i+1}_i \leftarrow \bm{y}^{k_i}_i + \alpha\frac{ d\mathcal{L}(\bm{y}_1^{k_1},...,\bm{y}_i^{k_i},\bm{y}^K_{>i})}{d \bm{y}^{k_i}_i},\notag\\\textrm{where }\bm{y}_{>i}^0=f_{\phi}(\bm{x},\bm{y}_1^{k_1},...,\bm{y}_i^{k_i}),\label{eq:incygw}
\end{gather}
where $\bm{y}_i^{k_i}$ denotes the latent $\bm{y}_i$ after $k_i$ steps of update. In next section, we show how to correct both of the above-mentioned issues by recursively applying back-propagating through gradient ascent \citep{domke2012generic}.
\subsection{An Accurate Implementation}
\label{sec:adv}
\textbf{Accurate SAVI on 2-level non-factorized latent}
We first extend the original SAVI on 1-level latent \citep{kim2018semi} to 2-level non-factorized latent. As the notation in NVC, we denote $\bm{x}$ as evidence, $\bm{y}_1$ as the variational posterior parameter of the first level latent $\bmt{y}_1$, $\bm{y}_2$ as the variational posterior parameter of the second level latent $\bmt{y}_2$, and the ELBO to maximize as $\mathcal{L}(\bm{y}_1,\bm{y}_2)$. The posterior $q(\bmt{y}_1,\bmt{y}_2|\bm{x})$ factorizes as $q(\bmt{y}_1|\bm{x})q(\bmt{y}_2|\bmt{y}_1,\bm{x})$, which means that $\bm{y}_2$ depends on $\bm{y}_1$. Given $\bm{y}_1$ is fixed, we can directly optimize $\bm{y}_2$ by gradient ascent. However, it requires some tricks to optimize $\bm{y}_1$.
The intuition is, we do not want to find a $\bm{y}_1$ that maximizes $\mathcal{L}(\bm{y}_1,\bm{y}_2)$ given a fixed $\bm{y}_2$. Instead, we want to find a $\bm{y}_1$, whose $\max_{\bm{y}_2}\mathcal{L}(\bm{y}_1,\bm{y}_2)$ is maximum. This translates to the optimization problem as:
\begin{gather}
    \bm{y}_1\leftarrow \arg \max_{\bm{y}_1} \mathcal{L}(\bm{y}_1,\bm{y}_2^*(\bm{y}_1)),\notag \\\textrm{ where } \bm{y}_2^*(\bm{y}_1)\leftarrow \arg\max_{\bm{y}_2} \mathcal{L}(\bm{y}_1,\bm{y}_2).
    \label{eq:opt2}
\end{gather}
In fact, Eq.~\ref{eq:opt2} is a variant of setup in back-propagating through gradient ascent \citep{samuel2009learning,domke2012generic}. The difference is, our $\bm{y}_1$ also contributes directly to optimization target $\mathcal{L}(\bm{y}_1,\bm{y}_2)$. From this perspective, Eq.~\ref{eq:opt2} is also closely connected to \citet{kim2018semi}, if we treat $\bm{y}_1$ as the amortized encoder parameter and $\bm{y}_2$ as latent. 

And as SAVI on 1-level latent \citep{kim2018semi}, we need to solve Eq.~\ref{eq:opt2} using gradient ascent. Specifically, denote $\alpha$ as learning rate, $K$ as the total gradient ascent steps, $\bm{y}_1^{k_1}$ as the $\bm{y}_1$ after $k_1$ step update, $\bm{y}_2^{k_2}$ as the $\bm{y}_2$ after $k_2$ step update, and $f(.)$ as FAVI initialing posterior parameters $\bm{y}_1^0,\bm{y}_2^0$, the optimization problem as Eq.~\ref{eq:opt2} translates into the update rule as: 
\begin{gather}
    \bm{y}_1^{k_1+1}\leftarrow \bm{y}_1^{k_1}+\alpha\frac{d\mathcal{L}(\bm{y}_1^{k_1},\bm{y}_2^{K})}{d\bm{y}_1^{k_1}},\notag\\
    \bm{y}_2^{k_2+1}\leftarrow \bm{y}_2^{k_2}+\alpha \frac{d\mathcal{L}(\bm{y}_1^{k_1},\bm{y}_2^{k_2})}{d\bm{y}_2^{k_2}}\textrm{, where } \bm{y}_2^0 = f(\bm{x},\bm{y}_1^{k_1}).\label{eq:opt2grad}
\end{gather}
To solve Eq.~\ref{eq:opt2grad}, we note that although $d\mathcal{L}(\bm{y}_1^{k_1},\bm{y}_2^{k_2})/d\bm{y}_2^{k_2}$ can be directly computed, $d\mathcal{L}(\bm{y}_1^{k_1},\bm{y}_2^{K})/d\bm{y}_1^{k_1}$ is not straightforward. Let's consider a simple example when the gradient ascent step $K=1$:
\begin{itemize}
    \item First, we initialize $\bm{y}_1,\bm{y}_2$ by FAVI $\bm{y}_1^0 \leftarrow \textrm{FAVI}(\bm{x}), \bm{y}_2^0 \leftarrow \textrm{FAVI}(\bm{x},\bm{y}_1^0)$.
    \item Next, we optimize $\bm{y}_2$ by one step gradient ascent as
    $\bm{y}_2^1 \leftarrow \bm{y}_2^0 + \alpha d \mathcal{L}(\bm{y}_2^0,\bm{y}_1^0) / d \bm{y}_2^0$ and evaluate ELBO as $\mathcal{L}(\bm{y}_1^0,\bm{y}_2^1)$.
    \item Next, we optimize $\bm{y}_1$ to maximize $\mathcal{L}(\bm{y}_1^0,\bm{y}_2^1)$, and the gradient is
    \begin{gather}
\frac{d \mathcal{L}(\bm{y}_1^0,\bm{y}_2^1)}{d \bm{y}_1} = \frac{\partial \mathcal{L}(\bm{y}_1^0,\bm{y}_2^1)}{\partial \bm{y}_1^0} + \frac{\partial \bm{y}_2^0}{\partial \bm{y}_1^0}\frac{d \mathcal{L}(\bm{y}_1^0,\bm{y}_2^1)}{d \bm{y}_2^0}.\notag
    \end{gather}
    With FAVI relationship, we naturally have $\partial \bm{y}_2^0 / \partial \bm{y}_1^0$. And we need to evaluate is $d \mathcal{L}(\bm{y}_1^0,\bm{y}_2^1) / d \bm{y}_2^0$. And this is where back-prop through gradient ascent \citep{samuel2009learning,domke2012generic} works:
    \begin{align}
    \frac{d \mathcal{L}(\bm{y}_1^0,\bm{y}_2^1)}{d \bm{y}_2^0} &= \frac{\partial \bm{y}_2^1}{\partial \bm{y}_2^0}\frac{d \mathcal{L}(\bm{y}_1^0,\bm{y}_2^1)}{d \bm{y}_2^1} \notag\\
    &= (I+\alpha\frac{\partial^2 \mathcal{L}(\bm{y}_2^0,\bm{y}_1^0)}{\partial^2 \bm{y}^0_2})\frac{d \mathcal{L}(\bm{y}_1^0,\bm{y}_2^1)}{d \bm{y}_2^1}. \notag
    \end{align}
    Again, $d \mathcal{L}(\bm{y}_1^0,\bm{y}_2^1)/d \bm{y}_2^1$ is known.
    \item By now, we have collect all parts to solve $d \mathcal{L}(\bm{y}_1^0,\bm{y}_2^1) / d \bm{y}_1^0$. And we can finally update $\bm{y}_1^0$ as $\bm{y}_1^1 \leftarrow \bm{y}_1^0 + \alpha d \mathcal{L}(\bm{y}_1^0,\bm{y}_2^1) / d \bm{y}_1^0$.
\end{itemize}

For $K>1$, we can extend the example and implement Eq.~\ref{eq:opt2grad} as Alg.~\ref{alg:advance2}. Specifically, we first initialize $\bm{y}_1^0$ from FAVI. Then we conduct gradient ascent on $\bm{y}_1$ with gradient $d\mathcal{L}(\bm{y}_1^{k_1},\bm{y}_2^K)/d\bm{y}_1^{k_1}$ computed from the procedure grad-2-level($\bm{x},\bm{y}_1^{k_1}$). And each time grad-2-level($\bm{x},\bm{y}_1^{k_1}$) is evaluated, $\bm{y}_2$ goes through a re-initialization and $K$ steps of gradient ascent. The above procedure corresponds to Eq.~\ref{eq:opt2grad}. The key of Alg.~\ref{alg:advance2} is the evaluation of gradient $d\mathcal{L}(\bm{a}^k,\bm{b}^K)/d\bm{a}^{k}$. Formally, we have:
\begin{theorem}
\label{th:2l}
After \textup{grad-2-level($\bm{x},\bm{y}_1^{k_1}$)} of Alg.~\ref{alg:advance2} executes, we have the return value $d \mathcal{L}(\bm{y}_1^{k_1},\bm{y}_2^K)/d\bm{y}_1^{k_1}=\overleftarrow{\bm{y}_1}$. (See proof in Appendix.~\ref{app:proof})
\end{theorem}
\begin{algorithm}[thb]
    \caption{Proposed Accurate SAVI on 2-level Latent}
    \label{alg:advance2}
\begin{algorithmic}
    \STATE {\bfseries procedure} solve-2-level($\bm{x}$)
    \STATE initialize $\bm{y}_1^0\leftarrow f_{\phi}(\bm{x})$ from FAVI.
    \FOR{$k_1=0$ {\bfseries to} $K-1$}
    \STATE $\frac{d\mathcal{L}(\bm{y}_1^{k_1},\bm{y}_2^K)}{d\bm{y}_1^{k_1}}=\textrm{grad-2-level}(\bm{x},\bm{y}_1^{k_1})$
    \STATE $\bm{y}_1^{k_1+1}\leftarrow\bm{y}_1^{k_1}+\alpha\frac{d\mathcal{L}(\bm{y}_1^{k_1},\bm{y}_2^K)}{d\bm{y}_1^{k_1}}$
    \ENDFOR
    \STATE {\bfseries return} $\bm{y}_1^K,\bm{y}_2^K$
    \STATE
    \STATE {\bfseries procedure} grad-2-level($\bm{x},\bm{y}_1^{k_1}$)
    \STATE initialize $\bm{y}_2^0\leftarrow f_{\phi}(\bm{x},\bm{y}_1^{k_1})$ from FAVI.
    \FOR{$k_2=0$ {\bfseries to} $K-1$}
    \STATE $\bm{y}_2^{k_2+1}\leftarrow\bm{y}_2^{k_2}+\alpha\frac{d\mathcal{L}(\bm{y}_1^{k_1},\bm{y}_2^{k_2})}{d\bm{y}_2^{k_2}}$
    \ENDFOR
    \STATE initialize $\overleftarrow{\bm{y}_1}\leftarrow\frac{\partial \mathcal{L}(\bm{y}_1^{k_1},\bm{y}_2^K)}{\partial \bm{y}_1^{k_1}}$, $\overleftarrow{\bm{y}_2}^K\leftarrow\frac{d\mathcal{L}(\bm{y}_1^{k_1},\bm{y}_2^K)}{d\bm{y}_2^K}$
    \FOR{$k_2=K-1$ {\bfseries to} $0$}
    \STATE $\overleftarrow{\bm{y}_1}\leftarrow\overleftarrow{\bm{y}_1}+\alpha\frac{\partial^2 \mathcal{L}(\bm{y}_1^{k_1},\bm{y}_2^{k_2})}{\partial \bm{y}_1^{k_1}\partial\bm{y}_2^{k_2}}\overleftarrow{\bm{y}_2}^{k_2+1}$
    \STATE $\overleftarrow{\bm{y}_2}^{k_2}\leftarrow\overleftarrow{\bm{y}_2}^{k_2}+\alpha\frac{\partial^2 \mathcal{L}(\bm{y}_1^{k_1},\bm{y}_2^{k_2})}{\partial \bm{y}_2^{k_2}\partial\bm{y}_2^{k_2}}\overleftarrow{\bm{y}_2}^{k_2+1}$
    \ENDFOR
    \STATE $\overleftarrow{\bm{y}_1}=\overleftarrow{\bm{y}_1}+\frac{\partial \bm{y}_2^0}{\partial\bm{y}_1^{k_1}}\overleftarrow{\bm{y}_2}^{0}$
    \STATE {\bfseries return} $\frac{d\mathcal{L}(\bm{y}_1^{k_1},\bm{y}_2^K)}{d\bm{y}_1^{k_1}}=\overleftarrow{\bm{y}_1}$
\end{algorithmic}
\end{algorithm}
\textbf{Accurate SAVI on DAG Latent} Then, we extend the result on 2-level latent to general non-factorized latent with dependency described by DAG. This DAG is the computational graph during network inference, and it is also the directed graphical model (DGM) \citep{koller2009probabilistic} defining the factorization of latent variables during inference. This extension is necessary for SAVI on latent with complicated dependency (e.g. bit allocation of NVC).

Similar to the 2-level latent setup, we consider performing SAVI on $N$ variational posterior parameter $\bm{y}_1,...,\bm{y}_N$ with their dependency defined by a computational graph $\mathcal{G}$, i.e., their corresponding latent variable $\bmt{y}_1,...,\bmt{y}_N$'s posterior distribution factorizes as $\mathcal{G}$. Specifically, we denote $\bm{y}_j\in\mathcal{C}(\bm{y}_i),\bm{y}_i\in\mathcal{P}(\bm{a}_j)$ if an edge exists from $\bm{y}_i$ to $\bm{y}_j$. This indicates that $\bmt{y}_j$ conditions on $\bmt{y}_i$. Without loss of generality, we assume $\bm{y}_1,...,\bm{y}_N$ is sorted in topological order. This means that if $\bm{y}_j\in\mathcal{C}(\bm{y}_i),\bm{y}_i\in\mathcal{P}(\bm{y}_j)$, then $i<j$. Each latent is optimized by $K$-step gradient ascent, and $\bm{y}_i^{k_i}$ denotes the latent $\bm{y}_i$ after $k_i$ steps of update. Then, similar to 2-level latent, we can solve this problem by recursively applying back-propagating through gradient 
ascent \citep{samuel2009learning,domke2012generic} to obtain Alg.~\ref{alg:advancedag}.

Specifically, we add a fake latent $\bm{y}_0$ to the front of all $\bm{y}$s. Its children are all the $\bm{y}$s with 0 in-degree. Then, we can solve the SAVI on $\bm{y}_1,...,\bm{y}_N$ using gradient ascent by executing the procedure grad-dag($\bm{x},\bm{y}_0^{k_0},...,\bm{y}_i^{k_i}$) in Alg.~\ref{alg:advancedag} recursively. Inside procedure grad-dag($\bm{x},\bm{y}_0^{k_0},...,\bm{y}_i^{k_i}$), the gradient to update $\bm{y}_i$ relies on the convergence of its children $\bm{y}_{j}\in\mathcal{C}(\bm{y}_i)$, which is implemented by the recursive depth-first search (DFS) in line 11. And upon the completion of procedure grad-dag($\bm{x},\bm{y}_0^0$), all the latent converges to $\bm{y}_1^K,...,\bm{y}_N^K$. Similar to the 2-level latent case, the key of Alg.~\ref{alg:advancedag} is the evaluation of gradient $d\mathcal{L}(\bm{y}_0^{k_0},...,\bm{y}_i^{k_i},\bm{y}_{>i}^K)/d \bm{y}_i^{k_i}$. Formally, we have:
\begin{theorem}
\label{th:dag}
After the procedure \textup{grad-dag($\bm{x},\bm{y}_0^{k_0},...,\bm{y}_i^{k_i}$)} in Alg.~\ref{alg:advancedag} executes, we have the return value $d\mathcal{L}(\bm{y}_0^{k_0},...,\bm{y}_i^{k_i},\bm{y}_{>i}^K)/d \bm{y}_i^{k_i}=\overleftarrow{\bm{y}_i}$. (See proof in Appendix.~\ref{app:proof}.)
\end{theorem}
\begin{algorithm}[thb]
    \caption{Proposed Accurate SAVI on DAG Latent}
    \label{alg:advancedag}
\begin{algorithmic}
    \STATE {\bfseries procedure} solve-dag($\bm{x}$)
    \FOR{$\bm{y}_j$ with parent $\mathcal{P}(\bm{y}_j)=\varnothing$}
    \STATE add $\bm{y}_j$ to fake node $\bm{y}_0$'s children $\mathcal{C}(\bm{y}_0)$
    \ENDFOR
    \STATE grad-dag($\bm{x},\bm{y}_0^0$)
    \STATE {\bfseries return} $\bm{y}_1^K,...,\bm{y}_N^K$
    \STATE
    \STATE {\bfseries procedure} grad-dag($\bm{x},\bm{y}_0^{k_0},...,\bm{y}_i^{k_i}$)
    \FOR{$\bm{y}_{j}\in\mathcal{C}(\bm{y}_i)$ in topological order}
    \STATE initialize $\bm{y}_{j}^0\leftarrow f(\bm{x},\bm{y}_0^{k_0},...,\bm{y}_{<j}^{k_{<j}})$ from SAVI
    \FOR{$k_j=0,...,K-1$}
    \STATE $\frac{d\mathcal{L}(\bm{y}_0^{k_0},...,\bm{y}_j^{k_j},\bm{y}_{>j}^K)}{d\bm{y}_{j}^{k_j}} \leftarrow \textrm{grad-dag}(\bm{x},\bm{y}_0^{k_0},...,\bm{y}_j^{k_j})$
    \STATE $\bm{y}_{j}^{k_j+1}\leftarrow\bm{y}_{j}^{k_j}+ \alpha\frac{d\mathcal{L}(\bm{y}_0^{k_0},...,\bm{y}_j^{k_j},\bm{y}_{>j}^K)}{d\bm{y}_{j}^{k_j}}$
    \ENDFOR
    \ENDFOR
    \STATE $\overleftarrow{\bm{y}_i}\leftarrow \frac{\partial\mathcal{L}(\bm{y}_0^{k_0},...,\bm{y}_i^{k_i},\bm{y}_{>i}^K)}{\partial \bm{y}_i^{k_i}}$
    \FOR{$\bm{y}_{j}\in\mathcal{C}(\bm{y}_i)$ in topological order} 
    \STATE $\overleftarrow{\bm{y}_{j}}^K\leftarrow\frac{d \mathcal{L}(\bm{y}_0^{k_0},...,\bm{y}_i^{k_i},\bm{y}_{>i}^K)}{d\bm{y}_{j}^K}$
    \FOR{$k_j=K-1,...,0$}
    \STATE $\overleftarrow{\bm{y}_i}\leftarrow\overleftarrow{\bm{y}_i}+\alpha \frac{\partial^2 \mathcal{L}(\bm{y}_0^{k_0},...,\bm{y}_j^{k_j},\bm{y}_{>j}^K)}{\partial\bm{y}_i^{k_i}\partial\bm{y}_{j}^{k_j}}\overleftarrow{\bm{y}_{j}}^{k_j+1}$
    \STATE $\overleftarrow{\bm{y}_j}^{k_j}\leftarrow\overleftarrow{\bm{y}_j}^{k_j+1}+\alpha \frac{\partial^2 \mathcal{L}(\bm{y}_0^{k_0},...,\bm{y}_j^{k_j},\bm{y}_{>j}^K)}{\partial\bm{y}_j^{k_j}\partial\bm{y}_{j}^{k_j}}\overleftarrow{\bm{y}_{j}}^{k_j+1}$
    \ENDFOR
    \STATE $\overleftarrow{\bm{y}_i}\leftarrow \overleftarrow{\bm{y}_i} +\frac{\partial \bm{y}_j^0}{\partial \bm{y}_i^{k_i}}\overleftarrow{\bm{y}_j}^0$
    \ENDFOR
    \STATE {\bfseries return} $\frac{d\mathcal{L}(\bm{y}_0^{k_0},...,\bm{y}_i^{k_i},\bm{y}_{>i}^K)}{d\bm{y}_{i}^{k_i}}=\overleftarrow{\bm{y}_i}$
\end{algorithmic}
\end{algorithm}

\section{Complexity Reduction} 
An evident problem of the accurate SAVI (Sec.~\ref{sec:adv}) is the temporal complexity. Given the frame number $N$ and gradient ascent step $K$, Alg.~\ref{alg:advancedag} has temporal complexity of $\Theta(K^N)$. NVC with GoP size $10$ has approximately $N=20$ latent, and the SAVI on neural image compression \citep{yang2020improving,gaoflexible} takes around $K=2000$ step to converge. For bit allocation, the complexity of Alg.~\ref{alg:advancedag} is $\approx 2000^{20}$, which is intractable.

\subsection{Temporal Complexity Reduction}
\label{sec:approx}
Therefore, we provide an approximation to the SAVI on DAG. The general idea is that, the SAVI on DAG (Alg.~\ref{alg:advancedag}) satisfies both requirement on gradient signal described in Sec.~\ref{sec:problem}. We can not make it tractable without breaking them. Thus, we break one of them for tractable complexity, while maintaining good performance. Specifically, We consider the approximation as:
\begin{align}
\frac{d\mathcal{L}(\bm{y}_0^{k_0},...,\bm{y}_i^{k_i},\bm{y}_{>i}^K)}{d \bm{y}_i^{k_i}}\approx\frac{d\mathcal{L}(\bm{y}_0^{k_0},...,\bm{y}_i^{k_i},\bm{y}_{>i}^0)}{d \bm{y}_i^{k_i}},
\label{eq:approx}
\end{align}
which obeys the requirement 1 in Sec.~\ref{sec:problem} while breaks the requirement 2. Based on Eq.~\ref{eq:approx}, we design an approximation of SAVI on DAG. Specifically, with the approximation in Eq.~\ref{eq:approx}, the recurrent gradient computation in Alg.~\ref{alg:advancedag} becomes unnecessary as the right hand side of Eq.~\ref{eq:approx} does not require $\bm{y}_{>i}^K$. However, to maintain the dependency of latent, we still need to ensure that the children node $\bm{y}_{j}\in \mathcal{C}(\bm{y}_i)$ are re-initialized by FAVI every-time when $\bm{y}_i$ is updated. Therefore, we traverse the graph in topological order, keep the children node $\bm{y}_j$ untouched until all its parent node $\bm{y}_i\in\mathcal{P}(\bm{y}_j)$'s gradient ascent is completed. And the resulting approximate SAVI is Alg.~\ref{alg:approx}. Its temporal complexity is $\Theta(KN)$.
\begin{algorithm}[thb]
    \caption{Proposed Approximated SAVI}
    \label{alg:approx}
\begin{algorithmic}
    \STATE {\bfseries procedure} solve-approx($\bm{x}$)
    \FOR{$i=1$ {\bfseries to} $N$}
    \STATE initialize $\bm{y}_i^0,...,\bm{y}_N^0\leftarrow f_{\phi}(\bm{x},\bm{y}_{<i}^K)$ from FAVI.
    \FOR{$k=0$ {\bfseries to} $K-1$}
    \STATE $\frac{d\mathcal{L}(\bm{y}_{<i}^K,\bm{y}_i^k,\bm{y}_{>i}^K)}{d\bm{y}_i^k}\approx\frac{d\mathcal{L}(\bm{y}_{<i}^K,\bm{y}_i^k,\bm{y}_{>i}^0)}{d\bm{y}_i^k}$
    \STATE $\bm{y}_i^{k+1}\leftarrow\bm{y}_i^k+\alpha\frac{d\mathcal{L}(\bm{y}_{<i}^K,\bm{y}_i^k,\bm{y}_{>i}^K)}{d\bm{y}_i^k}$
    \ENDFOR
    \ENDFOR
    \STATE {\bfseries return} $\bm{y}_1^K,...,\bm{y}_N^K$
\end{algorithmic}
\end{algorithm}

\subsection{Spatial Complexity Reduction}
\label{sec:scalable}
Despite the the approximated SAVI on DAG reduce the temporal complexity from $\Theta(K^N)$ to $\Theta(KN)$, the spatial complexity remains $\Theta(N)$. Though most of NVC approaches adopt a small GoP size of 10-12 \citep{lu2019dvc,li2021deep}, there are emerging approach extending GoP size to 100 \citep{hu2022coarse}. Therefore, it is important to reduce the spatial complexity to constant for scalability.

Specifically, our approximated SAVI on DAG uses the gradient of GoP-level likelihood $\mathcal{L}$, which takes $\Theta(N)$ memory. We empirically find that we can reduce the likelihood range for gradient evaluation from the whole GoP to $C$ future frames:
\begin{align}
\frac{d\mathcal{L}}{d \bm{y}_i^{k_i}}=\sum_{j=i}^{N}\frac{d\mathcal{L}_j}{d \bm{y}_i^{k_i}}\approx \sum_{j=i}^{min(i+C,N)}\frac{d\mathcal{L}_j}{d \bm{y}_i^{k_i}},
\label{eq:approx2}
\end{align}
where $C$ is a constant. Then, the spatial complexity is reduced to $\Theta(C)$, which is constant to GoP size $N$.
\section{Experimental Results}
\begin{table*}[thb]
\centering
\resizebox{\linewidth}{!}{
\begin{tabular}{@{}lllllllllll@{}}
\toprule
 & \multicolumn{5}{c}{BD-BR (\%) $\downarrow$} & \multicolumn{5}{c}{BD-PSNR (dB) $\uparrow$} \\ \cmidrule(rr){2-6}\cmidrule(ll){7-11}
Method & Class B & Class C & Class D & Class E & UVG & Class B &Class C & Class D & Class E & UVG \\ \midrule
\multicolumn{3}{@{}l@{}}{\textit{DVC \citep{lu2019dvc} as Baseline} } & & & & & &  \\
\citet{li2016lambda}$^*$ & 20.21 & 17.13 & 13.71 & 10.32 & 16.69 &-0.54&-&-&-0.32&-0.47 \\
\citet{li2022rate}$^*$ & -6.80 & -2.96 & 0.48 & -6.85 & -4.12 & 0.19&-&-&0.28 & 0.14 \\
OEU \citep{lu2020content} & -13.57 & -11.29 & -18.97 & -12.43 & -13.78 & 0.39 & 0.49 & 0.83 & 0.48 & 0.48 \\
Proposed-Scalable (Sec.~\ref{sec:scalable}) & -21.66 & -26.44 & -29.81 & -22.78 & -22.86 & 0.66 & 1.10 & 1.30 & 0.91 & 0.79 \\ 
Proposed-Approx (Sec.~\ref{sec:approx}) & -32.10 & -31.71 & -35.86 & -32.93 & -30.92 & 1.03 & 1.38 & 1.67 & 1.41 & 1.15 \\ 
\midrule
\multicolumn{3}{@{}l@{}}{\textit{DCVC \citep{li2021deep} as Baseline} } & & & & & &  \\
OEU \citep{lu2020content} & -10.75 & -14.34 & -16.30 & -7.15 & -16.07 & 0.30 & 0.58 & 0.74 & 0.29 & 0.50 \\
Proposed-Scalable (Sec.~\ref{sec:scalable}) & -24.67 & -24.71 & -24.71 & -30.35 & -29.68 & 0.65 & 0.95 & 1.12 & 1.15 & 0.83 \\
Proposed-Approx (Sec.~\ref{sec:approx}) & -32.89 & -33.10 & -32.01 & -36.88 & -39.66 & 0.91 & 1.37 & 1.55 & 1.58 & 1.18 \\ 
\midrule
\multicolumn{3}{@{}l@{}}{\textit{HSTEM \citep{10.1145/3503161.3547845} as Baseline} } & & & & & &  \\
Proposed-Scalable (Sec.~\ref{sec:scalable}) & -15.42 & -17.21 & -19.95 & -10.03 & -3.18 & 0.32 & 0.58 & 0.81 &  0.26 & 0.07 \\ \bottomrule
\end{tabular}
}
\caption{The BD-BR and BD-PSNR of our approach compared with baselines (w/o bit allocation) and other bit allocation approaches. $^*$ The data comes from \citet{li2022rate}. We mark some data with `-' as they are not reported by \citet{li2022rate}.}
\label{tab:taresult1}
\end{table*}
\subsection{Evaluation on Density Estimation}
As the proposed SAVI without approximation (Proposed-Accurate, Sec.~\ref{sec:adv}) is intractable for NVC, we evaluate it on small density estimation tasks. Specifically, we consider a 2-level VAE with inference model $\bm{x}\rightarrow\bmt{y}_1\rightarrow\bmt{y}_2$, where $\bm{x}$ is observed data, $\bmt{y}_1$ is the first level latent with dimension 100 and $\bmt{y}_2$ is the second level latent with dimension 50. We adopt the same 2-level VAE architecture as \citet{burda2015importance}. And the dataset we use is MNIST \citep{lecun1998gradient}. More details are provided in Appendix.~\ref{app:impl}.
\begin{table}[thb]
\centering
\resizebox{\columnwidth}{!}{
\begin{tabular}{@{}lll@{}}
\toprule
                      & NLL & t-test p-value\\ \midrule
FAVI (2-level VAE)               & $\le$ 98.3113 & -\\
Original SAVI               & $\le$ 91.5530 & $\le$ 0.001 (w/ FAVI)\\
Proposed-Approx (Sec.~\ref{sec:approx}) & $\le$ 91.5486 & $\le$ 0.001 (w/ Original)\\ 
Proposed-Accurate (Sec.~\ref{sec:adv}) & $\le$ 91.5482  & $\le$ 0.001 (w/ Approx)\\
\bottomrule
\end{tabular}
}

\caption{The NLL result on density estimation.}
\label{tab:savi}
\end{table}

We evaluate the negative log likelihood (NLL) lowerbound of FAVI, Original SAVI (Alg.\ref{alg:naive}), proposed SAVI without approximation (Proposed-Accurate, Sec.~\ref{sec:adv}) and proposed SAVI with approximation (Proposed-Approx, Sec.~\ref{sec:approx}). Tab.~\ref{tab:savi} shows that our Proposed-Accurate achieves a lower NLL ($\le 91.5482$) than original SAVI ($\le 91.5530$). And our Proposed-Approx ($\le 91.5486$) is only marginally outperformed by Proposed-Accurate, which indicates that this approximation does not harm performance significantly. As the mean NLL difference between Proposed-Accurate and Proposed-Approx is small, we additionally perform pair-wise t-test between methods and report p-values. And the results suggest that the difference between methods is significant ($p\le 0.001$).
\subsection{Evaluation on Bit Allocation}
\textbf{Experiment Setup} We evaluate our bit allocation method based on 3 NVC baselines: DVC \citep{lu2019dvc}, DCVC \citep{li2021deep} and HSTEM \citep{10.1145/3503161.3547845}. For all 3 baseline methods, we adopt the official pre-trained models. As DVC and DCVC do not provide I frame model, we adopt \citet{cheng2020learned} with pre-trained models provided by \citet{begaint2020compressai}. Following baselines, we adopt HEVC Common Testing Condition (CTC) \citep{bossen2013common} and UVG dataset \citep{mercat2020uvg}. And the GoP size is set to 10 for HEVC CTC and 12 for UVG dataset. The R-D performance is measured by Bjontegaard-Bitrate (BD-BR) and BD-PSNR \citep{bjontegaard2001calculation}. For the proposed approach, we evaluate the approximated SAVI (Alg.~\ref{alg:approx}) and its scalable version with $C=2$. We adopt Adam \citep{Kingma2014AdamAM} optimizer with $lr=1\times10^{-3}$ to optimize $\bm{y}_{1:N}$ for $K=2000$ iterations. For other bit allocation methods, we select a traditional $\lambda$-domain approach \citep{li2016lambda}, a recent $\lambda$-domain approach \citep{li2020rate} and online encoder update (OEU) \citep{lu2020content}. More details are presented in Appendix.~\ref{app:impl}.

\textbf{Main Results}
We extensively evaluate the R-D performance of proposed SAVI with approximation (Proposed-Approx, Sec.~\ref{sec:approx}) and proposed SAVI with scalable approximation (Proposed-Scalable, Sec.~\ref{sec:scalable}) on 3 baselines and 5 datasets. As Tab.~\ref{tab:taresult1} shows, for DVC \citep{lu2019dvc} and DCVC \citep{li2021deep}, our Proposed-Approx and Proposed-Scalable outperform all other bit allocation methods by large margin. It shows that the current best bit allocation method, OEU \citep{lu2020content}, still has a room of $0.5$ dB BD-PSNR to improve compared with our result. For HSTEM \citep{10.1145/3503161.3547845} with large model, we only evaluate our Proposed-Scalable. As neither OEU \citep{lu2020content} nor Proposed-Approx is able to run on HSTEM within 80G RAM limit. Moreover, as HSTEM has a bit allocation module, the effect of bit allocation is not as evident as DVC and DCVC. However, our bit allocation still brings a significant improvement of more than $0.4$ dB PSNR over HSTEM.
\begin{table}[thb]
\centering
\resizebox{\columnwidth}{!}{
\begin{tabular}{@{}llll@{}}
\toprule
                     & BD-BR (\%) & Enc Time (s) & RAM (GB) \\ \midrule
\multicolumn{3}{@{}l@{}}{\textit{DVC \citep{lu2019dvc} as Baseline}} & \\
Baseline             & - & $\le$1.0 & 1.02 \\
\citet{li2016lambda} & 13.71 & $\le$1.0 & 1.02 \\
\citet{li2022rate} & 0.48 & $\le$1.0 & 1.02 \\
OEU \citep{lu2020content}             & -18.97 & 15.2 & 5.83 \\
Original SAVI              & -14.76 & 158.5 & 3.05 \\
Original SAVI (per-frame)  & -20.12 & 55.3 & 2.94 \\
Proposed-Scalable (Sec.~\ref{sec:scalable}) & -29.81 & 74.6 & 2.12 \\
Proposed-Approx (Sec.~\ref{sec:approx})             & -35.86 &  528.3 & 5.46 \\\midrule
VTM 13.2 & - & 219.9 & 1.40 \\
\bottomrule
\end{tabular}
}
\caption{The R-D performance and the encoder resource consumption of different approaches. Encode time is per-frame and measured with AMD EPYC 7742 CPU and Nvidia A100 GPU. }
\label{tab:abl}
\end{table}

\begin{table}[thb]
\centering
\resizebox{\columnwidth}{!}{
\begin{tabular}{@{}llll@{}}
\toprule
                     & steps, lr & BD-BR (\%) & Enc Time (s) \\ \midrule
Proposed-Scalable (Sec.~\ref{sec:scalable}) & 2000, $1\times10^{-3}$ & -29.81 & 74.6  \\
& 1000, $2\times10^{-3}$ & -25.37 & 37.3 \\
& 500, $4\times10^{-3}$ & -18.99 & 18.6 \\
Proposed-Approx (Sec.~\ref{sec:approx}) &  2000, $1\times10^{-3}$ & -35.86 & 528.3  \\
 &  1000, $2\times10^{-3}$ & -31.21 & 264.1  \\
 &  500, $4\times10^{-3}$ & -25.60 & 132.0  \\
\bottomrule
\end{tabular}
}
\caption{The R-D performance and the encoding time consumption of different gradient ascent step and learning rate. We assume that the temporal complexity is approximately linear to steps.}
\label{tab:abl2}
\end{table}

\begin{figure}[thb]
\centering
    \begin{minipage}[t]{0.49\linewidth}
    \includegraphics[width=\linewidth]{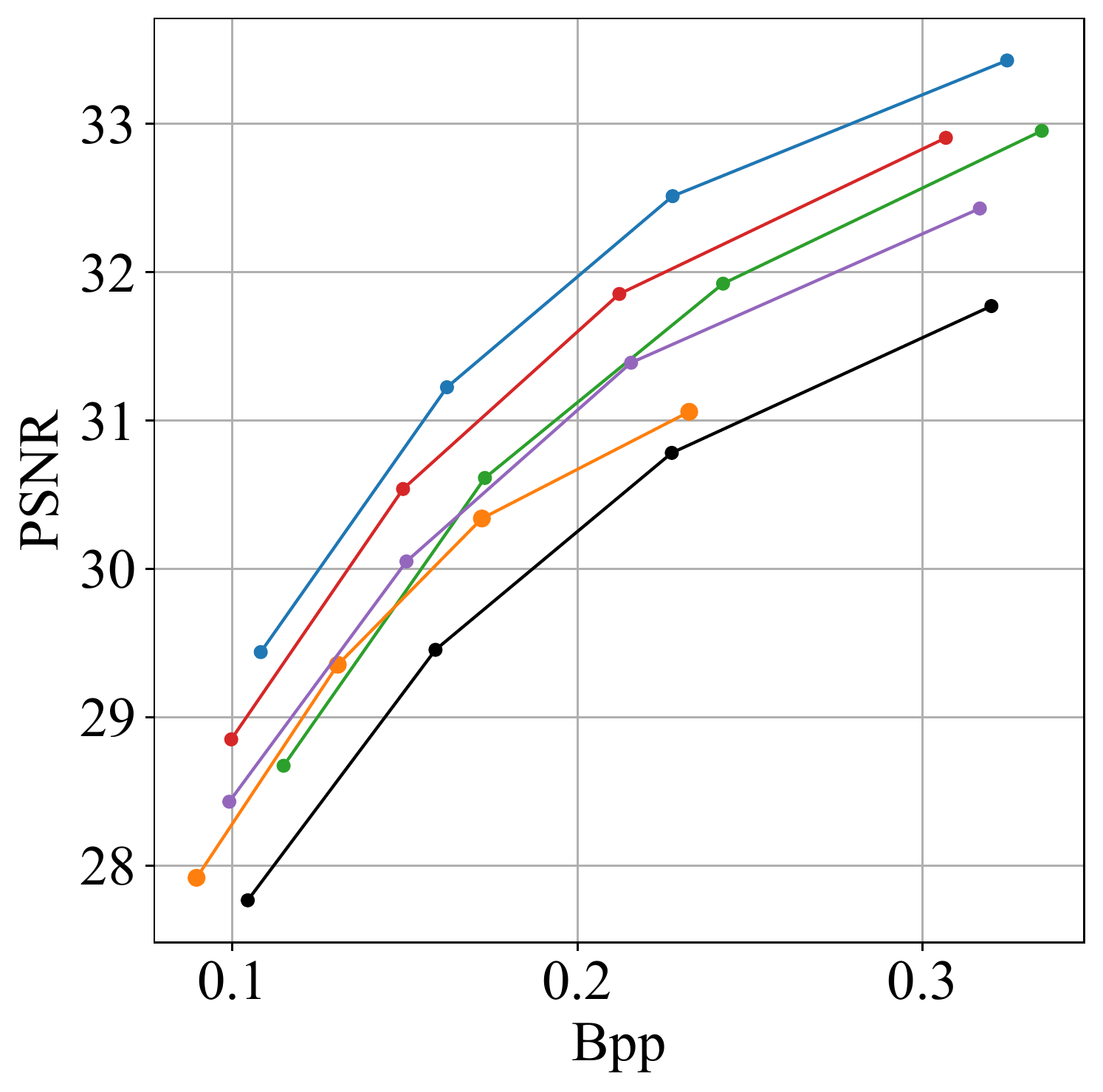}
    \end{minipage}%
    \hfill
    \begin{minipage}[t]{0.49\linewidth}
    \includegraphics[width=\linewidth]{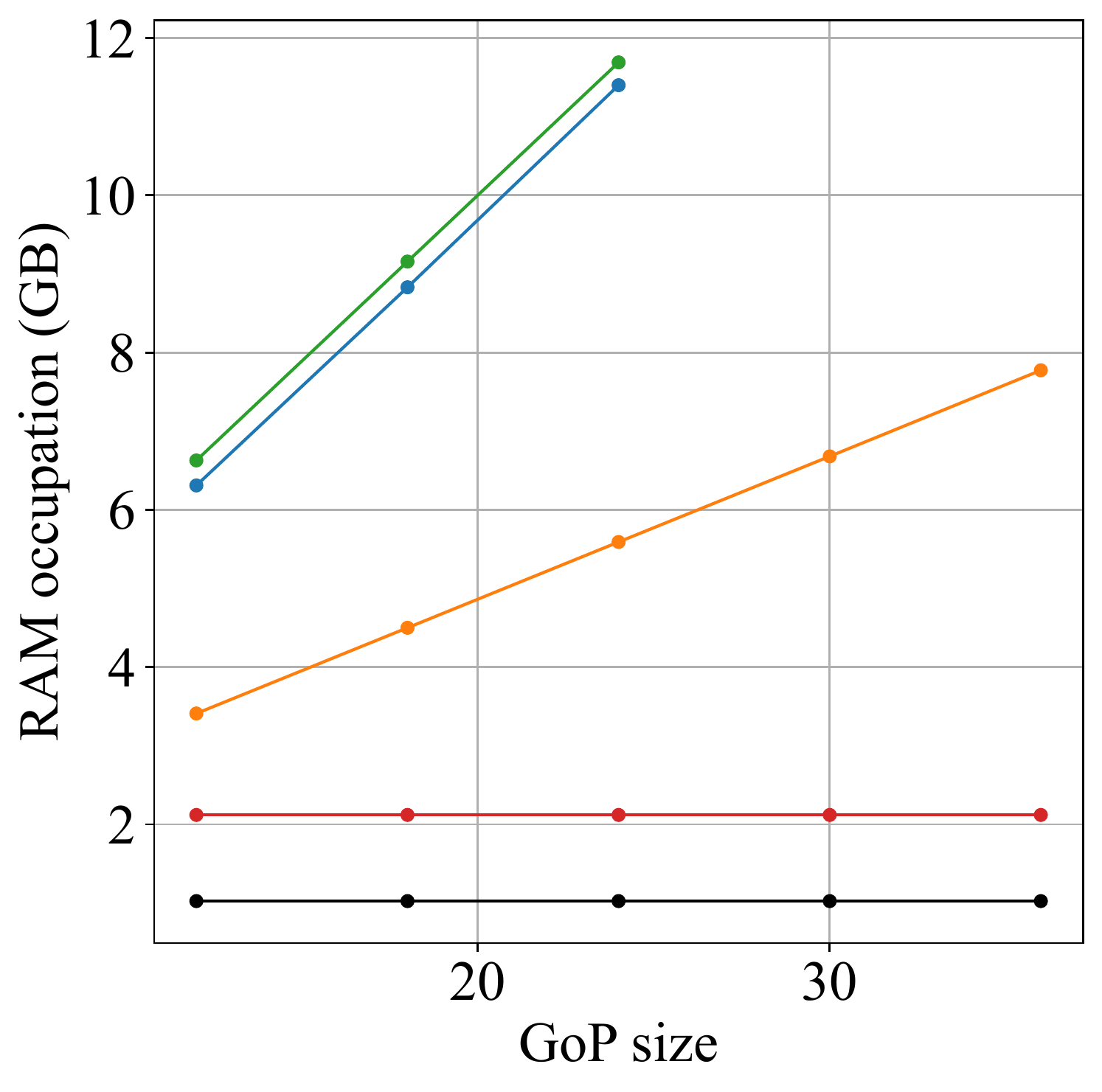}
    \end{minipage}%
    \hfill
    \begin{minipage}[t]{0.49\linewidth}
    \includegraphics[width=\linewidth]{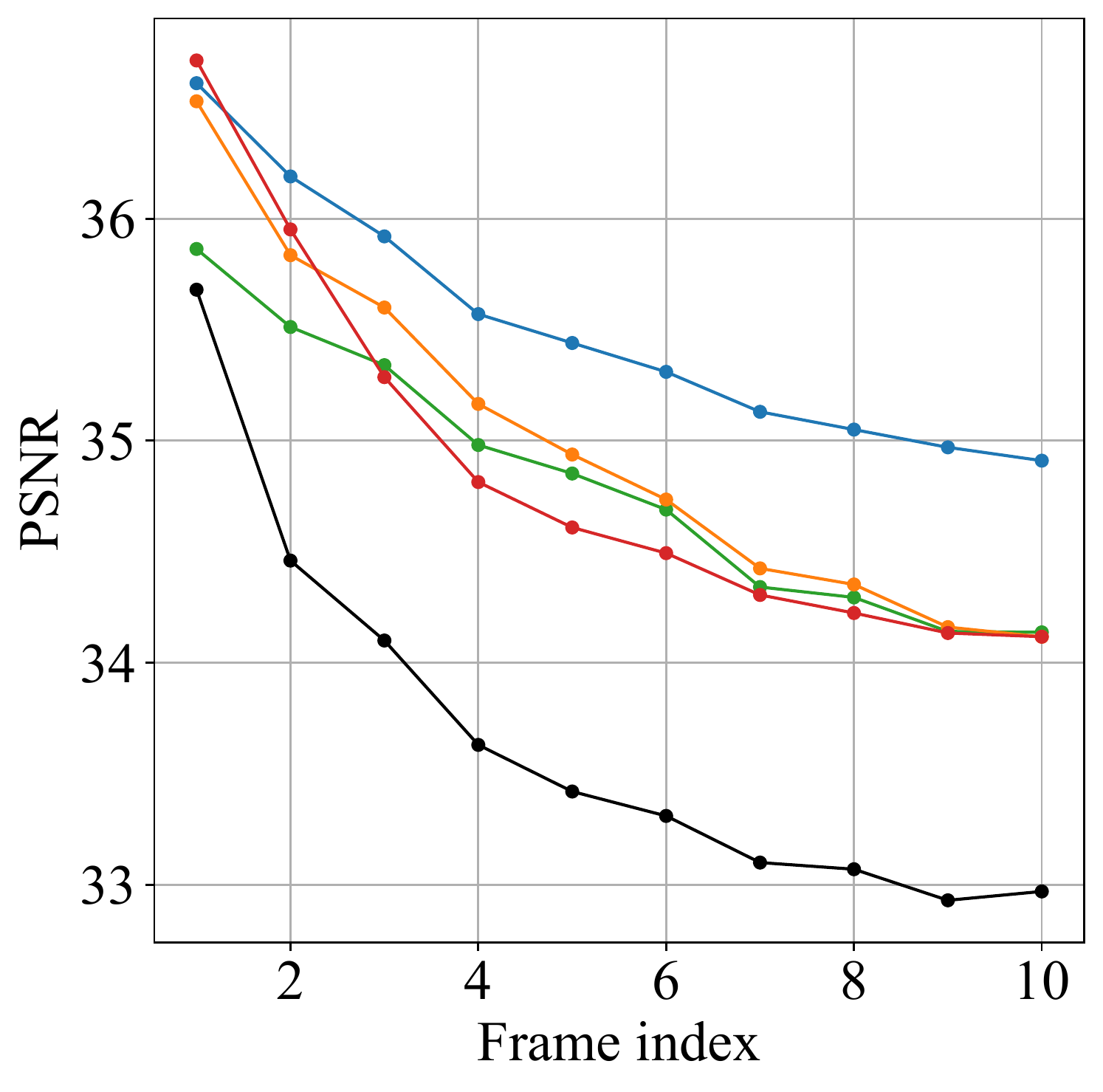}
    \end{minipage}%
    \hfill
    \begin{minipage}[t]{0.49\linewidth}
    \includegraphics[width=\linewidth]{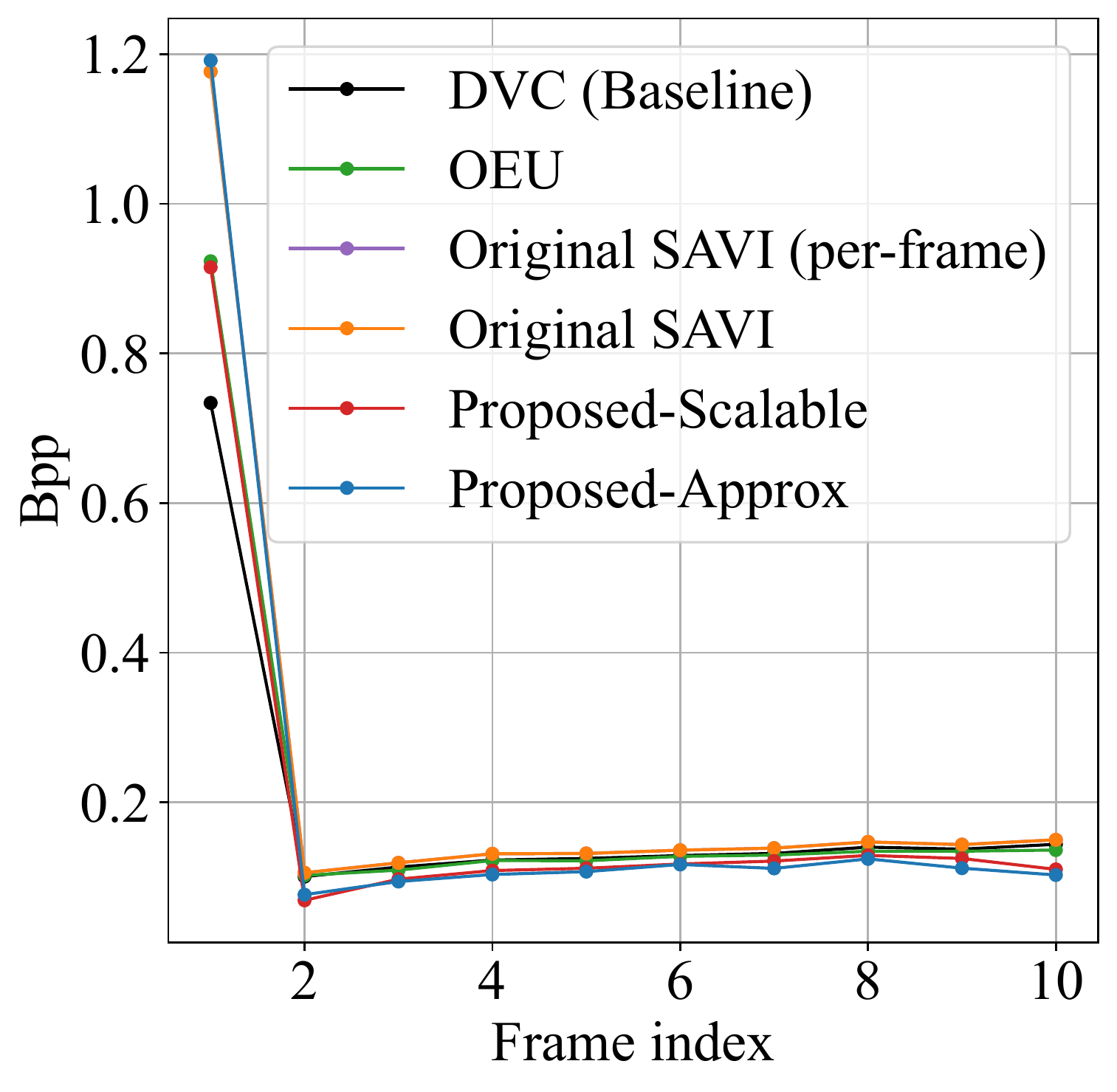}
    \end{minipage}%
\caption{\textit{upper-left.} The R-D performance of different methods. \textit{upper-right.} The RAM-GoP size relationship of different methods. \textit{lower-left.} The PSNR-frame index relationship before and after bit allocation. \textit{lower-right.} The bpp-frame index relationship before and after bit allocation.}
\label{fig:rdram}
\end{figure}

\textbf{Ablation Study}
We evaluate the R-D performance, encoding time and memory consumption of different methods with DVC \citep{lu2019dvc} on HEVC Class D. As Tab.~\ref{tab:abl} and Fig.~\ref{fig:rdram}.\textit{upper-left} show, the Proposed-Approx (-35.86\%) significantly outperforms the original SAVI (-14.76\%) in terms of BD-BR. Furthermore, it significantly outperforms all other bit allocation approaches. On the other hand, our Proposed-Scalable effectively decreases encoding time (from 528.3s to 74.6s). However, its R-D performance (-29.81\%) remains superior than all other bit allocation approaches by large margin. On the other hand, SAVI based approaches improve R-D performance not only by bit allocation, but also by reducing the amortization gap and discretization gap \citep{yang2020improving}. To investigate the net improvement by bit allocation, we perform \citep{yang2020improving} in a per-frame manner. And the resulting method (Original SAVI (per-frame)) improves the R-D performance by $20\%$. This means that the net enhancement brought by bit allocation is around $15\%$. Furthermore, as shown in Fig.~\ref{fig:rdram}.\textit{upper-right}, the memory requirement of our Proposed-Scalable is constant to GoP size. While for OEU \citep{lu2020content}, original SAVI and Proposed-Approx, the RAM is linear to GoP size. Despite the encoding time and RAM of our approach is higher than baseline, the decoding remains the same. Furthermore, compared with other encoder for R-D performance benchmark purpose (VTM 13.2), the encoding time and memory of our approach is reasonable. 

For both Proposed-Approx and Proposed-Scalable, it is possible to achieve speed-performance trade-off by adjusting gradient ascent steps and learning. In Tab.~\ref{tab:abl2}, we can see that reducing number of gradient ascent steps linearly reduce the encoding time while maintain competitive R-D performance. Specifically, Proposed-Scalable with $500$ steps and $1\times10^{-3}$ lr achieves similar BD-BR (-18.99 vs -18.97\%) and encoding time (18.6 vs 15.2s) as OEU \citep{lu2020content}, with less than half RAM requirement (2.12 vs 5.83GB).

\textbf{Analysis}
To better understand our bit allocation methods, we show the per-frame bpp and PSNR before and after bit allocation in Fig.~\ref{fig:rdram}.\textit{lower-left} and Fig.~\ref{fig:rdram}.\textit{lower-right}. It can be seen that the frame quality of baseline method drops rapidly during encoding process (from $35.5$ to $33.0$ dB), while our proposed bit allocation approach alleviate this degradation (from $36.5$ to $35.0$ dB). On the other hand, all bit allocation methods allocates more bitrate to I frame. The original SAVI need more bitrate to achieve the same frame quality as the proposed approach, that is why its R-D performance is inferior.

\textbf{Qualitative Results} See Appendix.~\ref{app:exp}.

\section{Related Works}
\textbf{Bit Allocation for Neural Video Compression} \citet{rippel2019learned,li2022rate} are the pioneer of bit allocation for NVC, who reuse the empirical model from traditional codec. More recently, \citet{10.1145/3503161.3547845} propose a feed-forward bit allocation approach with quantization step-size predicted by neural network. Other works only adopt simple heuristic such as inserting $1$ high quality frame per $4$ frames \citep{hu2022coarse,cetin2022flexible,li2023neural}. On the other hand, we also recognise OEU \citep{lu2020content} as frame-level bit allocation, while it follows the encoder overfitting proposed by \citet{cremer2018inference} instead of SAVI.

\textbf{Semi-Amortized Variational Inference for Neural Compression} SAVI is proposed by \citet{kim2018semi,marino2018iterative}. The idea is that works following \citet{kingma2013auto} use fully amortized inference parameter $\phi$ for all data, which leads to the amortization gap \citep{cremer2018inference}. SAVI reduces this gap by optimizing the variational posterior parameter after initializing it with inference network. It adopts back-propagating through gradient ascent \citep{domke2012generic} to evaluate the gradient of amortized parameters. When applying SAVI to practical neural codec, researchers abandon the nested parameter update for 
efficiency. Prior works \citep{djelouah2019content,yang2020improving,zhao2021universal,gaoflexible} adopt SAVI to boost R-D performance and achieve variable bitrate in image compression.

\section{Discussion \& Conclusion}
Despite the proposed approach has reasonable complexity as a benchmark, its encoding speed remains far from real-time. We will continue to investigate faster approximation. (See more in Appendix.~\ref{sec:disc}). Another limitation is that current evaluation is limited to NVC with P-frames.

To conclude, we show that the SAVI using GoP-level likelihood is equivalent to optimal bit allocation for NVC. We extent the original SAVI to non-factorized latent by back-propagating through gradient ascent and propose a feasible approximation to make it tractable for bit allocation. Experimental results show that current bit allocation methods still have a room of $0.5$ dB PSNR to improve, compared with our optimal result.

\section*{Acknowledgements}
This work was supported by Xiaomi AI Innovation Research under grant No.202-422-002.

\bibliography{example_paper}
\bibliographystyle{icml2022}

\clearpage
\newpage
\appendix
\section{Details of $\lambda$-domain Bit Allocation.}
\label{app:lamb}
The $\lambda$-domain bit allocation \citep{li2016lambda} is a well-acknowledged work in bit allocation. It is adopted in the official reference software of H.265 as the default bit allocation algorithm. In Sec.~\ref{sec:lam}, we briefly list the main result of \citet{li2016lambda} without much derivation. In this section, we dive into the detail of it.

To derive Eq.~\ref{eq:lamcon}, we note that we can expand the first optimal condition Eq.~\ref{eq:rdgrad} as:
\begin{align}
    \underbrace{\sum_{j=i}^N\frac{d R_j}{d R_i}}_{\textrm{Eq.~\ref{eq:dep}}}\frac{d R_i}{d\bm{\lambda}_i}+\bm{\lambda}_0^T \underbrace{\sum_{j=i}^N\frac{d \bm{D}_j}{d \bm{D}_i}}_{\textrm{Eq.~\ref{eq:dep}}}\frac{d \bm{D}_i}{d \bm{\lambda}_i} = 0.\label{eq:e11d1}
\end{align}
Then immediately we notice that the $dR_j/dR_i,d\bm{D}_j/d\bm{D}_i$ terms can be represented as the empirical model in Eq.~\ref{eq:dep}. After inserting Eq.~\ref{eq:dep}, we have:
\begin{align}
    I\frac{d R_i}{d\bm{\lambda}_i}+\bm{\lambda}_0^T \omega_iI\frac{d \bm{D}_i}{d \bm{\lambda}_i} = 0. \label{eq:e11d2}
\end{align}
And now we can obtain Eq.~\ref{eq:lamcon} by omitting the identity matrix $I$.

To derive Eq.~\ref{eq:lamlam}, we multiply both side of Eq.~\ref{eq:lamcon} by $d\bm{\lambda}_i/d\bm{D}_i$, and take the second optimal condition Eq.~\ref{eq:rdgradl} into it:
\begin{align}
    &\frac{d R_i}{d\bm{\lambda}_i}\frac{d\bm{\lambda}_i}{d\bm{D}_i}+\omega_i\bm{\lambda}_0^T\frac{d \bm{D}_i}{d \bm{\lambda}_i}\frac{d\bm{\lambda}_i}{d\bm{D}_i} \notag\\
    &=\underbrace{\frac{d R_i}{d\bm{D}_i}}_{\textrm{Eq.~\ref{eq:rdgradl}}}+\omega_i\bm{\lambda}_0^TI \notag\\
    &=-\bm{\lambda}^T_i + \omega_i\bm{\lambda}_0^TI \notag\\
    &=0. \label{eq:e12d}
\end{align}
And we can easily obtain Eq.~\ref{eq:lamlam} from the last lines of Eq.~\ref{eq:e12d}.

To see why our SAVI-based bit allocation generalizes $\lambda$-domain bit allocation, we can insert the $\lambda$-domain rate \& quality dependency model in Eq.~\ref{eq:dep} to the equivalent bit allocation map in Theorem.~\ref{th:eqls}:
\begin{align}
    \bm{\lambda}^{'}_i=&(I+\sum_{j=i+1}^{N}\frac{d \bm{D}_j}{d \bm{D}_{i}})^T\bm{\lambda}_0/(1+\sum_{j=i+1}^{N}\frac{d R_j}{d R_i}), \notag \\
    =& (\omega_i)^T\bm{\lambda}_0/(1+0),\notag \\
    =& \omega_i\bm{\lambda}_0,
\end{align}
and find out that we can obtain the result of $\lambda$-domain bit allocation. This implies that our SAVI-based bit allocation is the $\lambda$-domain bit allocation with a different rate \& quality dependency model, which is accurately defined via gradient of NVC, instead of empirical.
\section{Proof of Main Results.}
\label{app:proof}
In Appendix.~\ref{app:proof}, we provide proof to our main theoretical results.

\textbf{Theorem 3.1.}\textit{
The SAVI using GoP-level likelihood is equivalent to bit allocation with:
\begin{align}
    \bm{\lambda}^{'}_i=&(1+\sum_{j=i+1}^{N}\frac{d \bm{D}_j}{d \bm{D}_{i}})^T\bm{\lambda}_0/(1+\sum_{j=i+1}^{N}\frac{d R_j}{d R_i}), \tag{14}
\end{align}
where $d \bm{D}_j/d \bm{D}_{i},d R_j/d R_{i}$ can be computed numerically through the gradient of NVC model.
}
\begin{proof}
To find $\bm{\lambda}_i^{'}$, we expand the definition of $\bm{\lambda}_i^{'}$ in Eq.~\ref{eq:cee7} as:
\begin{align}
    &\frac{d (R_i+\bm{\lambda}^{'T}_{i}\bm{D}_i)}{d \bm{y}_i} \notag\\&= \frac{d (R_i+\bm{\lambda}^{T}_{0}\bm{D}_i)}{d \bm{y}_i} + \sum_{j=i+1}^{N}\frac{d (R_j+\bm{\lambda}^{T}_{0}\bm{D}_j)}{d \bm{y}_i}\notag\\&=(1+\sum_{j=i+1}^N\frac{d R_j}{d R_i})\frac{d R_i}{d \bm{y}_i}+\bm{\lambda}_0^{T}(1+\sum_{j=i+1}^N\frac{d \bm{D}_j}{d \bm{D}_i})\frac{d \bm{D}_i}{d \bm{y}_i}\notag\\&\propto d(R_i+\underbrace{\frac{(1+\sum_{j=i+1}^{N}\frac{d \bm{D}_j}{d \bm{D}_{i}})}{(1+\sum_{j=i+1}^{N}\frac{d R_j}{d R_i})}\bm{\lambda}_0^T}_{\bm{\lambda}_i^{'T}}\bm{D}_i)/d\bm{y}_i,\label{eq:cee8}
\end{align}
which implies that we have:
\begin{align}
    \bm{\lambda}^{'}_i=&(1+\sum_{j=i+1}^{N}\frac{d \bm{D}_j}{d \bm{D}_{i}})^T\bm{\lambda}_0/(1+\sum_{j=i+1}^{N}\frac{d R_j}{d R_i}). \notag
\end{align}
To evaluate the rate \& quality dependency model $d \bm{D}_j/d \bm{D}_i$ and $d R_j/d R_i$ numerically, we note that we can expand them as:
\begin{align}
    \frac{d R_j}{d R_i}=\frac{d R_j}{d\bm{y}_i}\underbrace{\frac{d \bm{y}_i}{d R_i}}_{\textrm{Eq.~}\ref{eq:nrd3}} \textrm{, } \frac{d \bm{D}_j}{d \bm{D}_i}=\frac{d \bm{D}_j}{d\bm{y}_i}\underbrace{\frac{d \bm{y}_i}{d \bm{D}_i}}_{\textrm{Eq.~}\ref{eq:nrd3}}.\label{eq:depm}
\end{align}
We note that $d R_j/d \bm{y}_i$ and $d \bm{D}_j/d \bm{y}_i$ can be evaluated directly on a computational graph. On the other hand, $d R_i/d \bm{y}_i$ and $d \bm{D}_i/d \bm{y}_i$ can also be evaluated directly on a computational graph. Then, we can numerically obtain the reverse direction gradient $d \bm{y}_i/d R_i$ and $d \bm{y}_i / d \bm{D}_i$ by taking reciprocal of each element of the Jacobian:
\begin{align}
(\frac{d \bm{y}_{i}}{d R_{i}})^{mn}=1/(\frac{d R_{i}}{d \bm{y}_{i}})^{nm} \textrm{, } (\frac{d \bm{y}_{i}}{d \bm{D}_{i}})^{mn}=1/(\frac{d\bm{D}_{i}}{d \bm{y}_{i}})^{nm},
    \label{eq:nrd3}
\end{align}
where $m,n$ are the location index of the Jacobian matrix. Now all the values in Eq.~\ref{eq:depm} are solved, we can numerically evaluate the rate \& quality dependency model.
\end{proof}

\textbf{Theorem 3.2.}\textit{
The equivalent bit allocation map $\bm{\lambda}^{'}_{i}$ is the solution to the optimal bit allocation problem in Eq.~\ref{eq:rd1}. In other words, we have:
\begin{align}
    \bm{\lambda}_i^{'} = \bm{\lambda}_i^*.\tag{15}
\end{align}
}
\begin{proof}
First, we will solve $\bm{\lambda}_i^*$ of optimal bit allocation in Eq.~\ref{eq:rd1} analytically with the precise R-D model defined by the computational graph. And then we will prove Theorem.~\ref{th:eqrd} by showing that $\bm{\lambda}_i^*=\bm{\lambda}_i^{'}$. 

To solve $\bm{\lambda}_i^{*}$, we expand $d R_j/d \bm{\lambda}_i$ and  $d \bm{D}_j/d\bm{\lambda}_i$ as:
\begin{gather}
    \frac{d R_j}{d \bm{\lambda}_i}=\frac{d R_j}{d R_i}\frac{d R_i}{d \bm{\lambda}_i}\textrm{, }\frac{d \bm{D}_j}{d \bm{\lambda}_i}=\frac{d \bm{D}_j}{d \bm{D}_i}\frac{d \bm{D}_i}{d \bm{\lambda}_i}.\label{eq:a3:p0}
\end{gather}
Then by taking Eq.~\ref{eq:a3:p0} into Eq.~\ref{eq:rdgrad}, we obtain:
\begin{align}
    (1+\sum_{j=i+1}^{N}\frac{d R_j}{d R_{i}})\frac{d R_i}{d\bm{\lambda}_i}+\bm{\lambda}_0^T(1+\sum_{j=i+1}^{N}\frac{d\bm{D}_j}{d\bm{D}_i})\frac{d\bm{D}_i}{d\bm{\lambda}_i}=0. \label{eq:a3:p1}
\end{align}
Further, multiply each side of Eq.~\ref{eq:a3:p1} by $d \bm{\lambda}_i/d \bm{D}_i$ and take Eq.~\ref{eq:rdgradl} into it, we have:
\begin{align}
    -\bm{\lambda}_i^{*T}(1+\sum_{j=i+1}^{N}\frac{d R_j}{d R_i})+\bm{\lambda}_0^T(1+\sum_{j=i+1}^{N}\frac{d \bm{D}_j}{d \bm{D}_i})=0. \label{eq:a3:p2}
\end{align}
Then immediately, we have:
\begin{align}
    \bm{\lambda}_i^*=&(1+\sum_{j=i+1}^{N}\frac{d\bm{D}_j}{d\bm{D}_i})^T\bm{\lambda}_0/(1+\sum_{j=i+1}^{N}\frac{d R_j}{d R_i})\notag\\=&\bm{\lambda}_i^{'}, \label{eq:a3:p3}
\end{align}
which proves Theorem.~\ref{th:eqrd}.
\end{proof}

\textbf{Theorem 3.3.}\textit{
After \textup{grad-2-level($\bm{x},\bm{y}_1^{k_1}$)} of Alg.~\ref{alg:advance2} executes, we have the return value $d \mathcal{L}(\bm{y}_1^{k_1},\bm{y}_2^K)/d\bm{y}_1^{k_1}=\overleftarrow{\bm{y}_1}$.
}
\begin{proof}
This proof extends the proof of Theorem.~1 in \citet{domke2012generic}. Note that our algorithm is different from \citet{samuel2009learning,domke2012generic,kim2018semi} as our high level parameter $\bm{y}_1$ not only generate low level parameter $\bm{y}_2$, but also directly contributes to optimization target (See Fig.~\ref{fig:cg2}).
\begin{figure}[htb]
\centering
    \includegraphics[width=0.6 \linewidth]{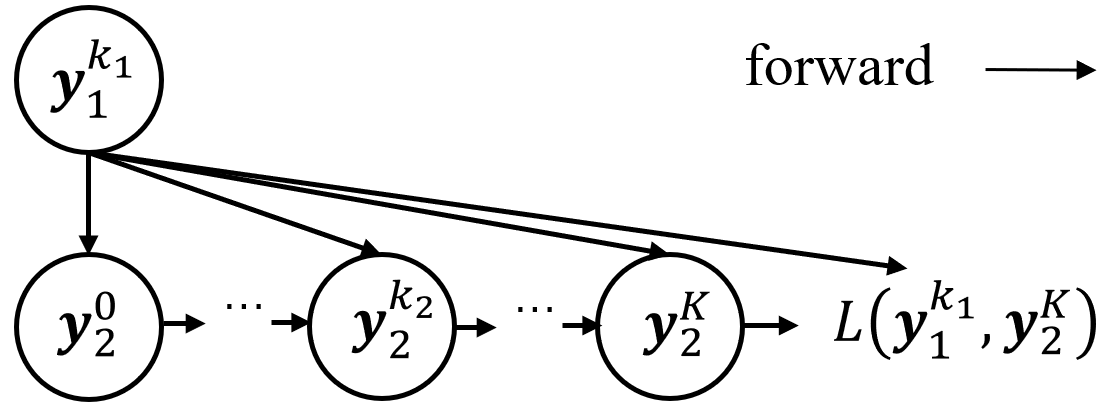}
    \caption{The computational graph corresponding to Eq.~\ref{eq:wlgd_gd}.}
    \label{fig:cg2}
\end{figure}

As the computational graph in Fig.~\ref{fig:cg2} shows, we can expand $d \mathcal{L}(\bm{y}_1^{k_1},\bm{y}_2^{K})/d\bm{y}_1^{k_1}$ as:
\begin{align}
    \frac{d \mathcal{L}(\bm{y}_1^{k_1},\bm{y}_2^{K})}{d\bm{y}_1^{k_1}}=\underbrace{\frac{\partial \mathcal{L}(\bm{y}_1^{k_1},\bm{y}_2^{K})}{\partial \bm{y}_1^{k_1}}}_{\textrm{known}}+\sum_{k_2=0}^{K}\underbrace{\frac{\partial \bm{y}_2^{k_2}}{\partial\bm{y}_1^{k_1}}}_{\textrm{Eq.~}\ref{eq:yw_gd}}\underbrace{\frac{d\mathcal{L}(\bm{y}_1^{k_1},\bm{y}_2^K)}{d\bm{y}_2^{k_2}}}_{\textrm{Eq.~}\ref{eq:yKyk_gd}}.
    \label{eq:wlgd_gd}
\end{align}
To solve Eq.~\ref{eq:wlgd_gd}, we first note that $\partial\mathcal{L}(\bm{y}_1^{k_1},\bm{y}_2^K)/\partial\bm{y}_1^{k_1}$, $d\mathcal{L}(\bm{y}_1^{k_1},\bm{y}_2^K)/d\bm{y}_2^K$ and $\partial\bm{y}_2^0/\partial\bm{y}_1^{k_1}$ is naturally known. Then, by taking partial derivative of the update rule of gradient ascent $\bm{y}_2^{k_2+1}\leftarrow \bm{y}_2^{k_2}+\alpha d\mathcal{L}(\bm{y}_1^{k_1},\bm{y}_2^{k_2})/d\bm{y}_2^{k_2}$ with regard to $\bm{y}_1^{k_1},\bm{y}_2^{k_2}$, we have:
\begin{gather}
    \frac{\partial \bm{y}_2^{k_2+1}}{\partial\bm{y}_2^{k_2}}=I+\alpha\frac{\partial^2\mathcal{L}(\bm{y}_1^{k_1},\bm{y}_2^{k_2})}{\partial\bm{y}_2^{k_2}\partial\bm{y}_2^{k_2}}\label{eq:yy_gd},\\
    \frac{\partial \bm{y}_2^{k_2+1}}{\partial\bm{y}_1^{k_1}}=\alpha\frac{\partial^2\mathcal{L}(\bm{y}_1^{k_1},\bm{y}_2^{k_2})}{\partial\bm{y}_1^{k_1}\partial\bm{y}_2^{k_2}}.
    \label{eq:yw_gd}
\end{gather}
Note that Eq.~\ref{eq:yw_gd} is the partial derivative $\partial \bm{y}_2^{k_2+1}/\partial\bm{y}_1^{k_1}$ instead of total derivative $d\bm{y}_2^{k_2+1}/d\bm{y}_1^{k_1}$, whose value is $(\partial \bm{y}_2^{k_2+1}/\partial\bm{y}_2^{k_2}) (d\bm{y}_2^{k_2}/d\bm{y}_1^{k_1})+\partial \bm{y}_2^{k_2+1}/\partial\bm{y}_1^{k_1}$.
And those second order terms can either be directly evaluated or approximated via finite difference as Eq.~\ref{eq:hes}.

As Eq.~\ref{eq:yw_gd} already solves the first term on the right hand side of Eq.~\ref{eq:wlgd_gd}, the remaining issue is $d\mathcal{L}(\bm{y}_1^{k_1},\bm{y}_2^K)/d\bm{y}_2^{k_2}$. To solve this term, we expand it recursively as:
\begin{align}
    \frac{d\mathcal{L}(\bm{y}_1^{k_1},\bm{y}_2^K)}{d\bm{y}_2^{k_2}}=\underbrace{\frac{\partial \bm{y}_2^{k_2+1}}{\partial\bm{y}_2^{k_2}}}_{\textrm{Eq.~}\ref{eq:yy_gd}}\frac{d\mathcal{L}(\bm{y}_1^{k_1},\bm{y}_2^K)}{d\bm{y}_2^{k_2+1}}.
    \label{eq:yKyk_gd}
\end{align}
Then, we have solved all parts that is required to evaluate $\mathcal{L}(\bm{y}_1^{k_1},\bm{y}_2^K)/d\bm{y}_1^{k_1}$. The above solving process can be described by the procedure grad-2-level($\bm{x},\bm{y}_1^{k_1}$) of Alg.~\ref{alg:advance2}. Specifically, the iterative update of $\overleftarrow{\bm{y}_2}^{k_2+1}$ corresponds to recursively expanding Eq.~\ref{eq:yKyk_gd} with Eq.~\ref{eq:yy_gd}, and the iterative update of $\overleftarrow{\bm{y}_1}$ corresponds to recursively expanding Eq.~\ref{eq:wlgd_gd} with Eq.~\ref{eq:yw_gd} and Eq.~\ref{eq:yKyk_gd}. Upon the return of grad-2-level($\bm{x},\bm{y}_1^{k_1}$) of Alg.~\ref{alg:advance2}, we have $\overleftarrow{\bm{y}_1}=d \mathcal{L}(\bm{y}_1^{k_1},\bm{y}_2^{K})/d\bm{y}_1^{k_1}$.

The complexity of the Hessian-vector product in Alg.~\ref{alg:advance2} may be reduced using finite difference following \citep{domke2012generic} as:
\begin{align}
    \frac{\partial^2 \mathcal{L}(\bm{y}_1^{k_1},\bm{y}_2^{k_2})}{\partial \bm{y}_1^{k_1}\partial\bm{y}_2^{k_2}}\bm{v}=&\notag\\\lim_{r\rightarrow 0}\frac{1}{r}(&\frac{d \mathcal{L}(\bm{y}_1^{k_1},\bm{y}_2^{k_2}+r\bm{v})}{d \bm{y}_1^{k_1}}-\frac{d \mathcal{L}(\bm{y}_1^{k_1},\bm{y}_2^{k_2})}{d \bm{y}_1^{k_1}}),\notag\\
    \frac{\partial^2 \mathcal{L}(\bm{y}_1^{k_1},\bm{y}_2^{k_2})}{\partial \bm{y}_2^{k_2}\partial\bm{y}_2^{k_2}}\bm{v}=&\notag\\\lim_{r\rightarrow 0}\frac{1}{r}(&\frac{d \mathcal{L}(\bm{y}_1^{k_1},\bm{y}_2^{k_2}+r\bm{v})}{d \bm{y}_2^{k_2}}-\frac{d \mathcal{L}(\bm{y}_1^{k_1},\bm{y}_2^{k_2})}{d \bm{y}_2^{k_2}}).\label{eq:hes}
\end{align}
\end{proof}

\begin{figure*}[thb]
\centering
\includegraphics[width=0.8\linewidth]{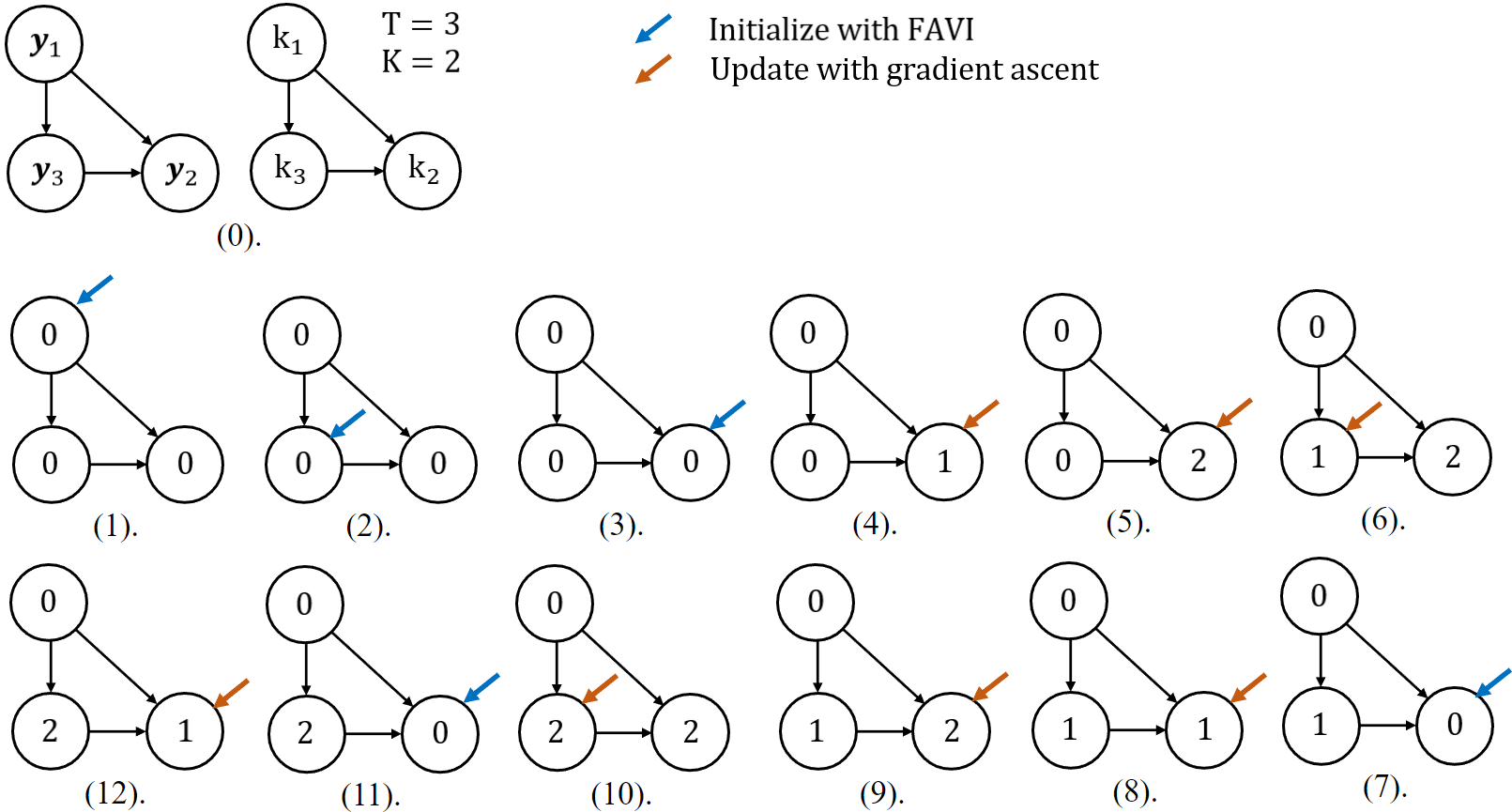}
\includegraphics[width=0.8\linewidth]{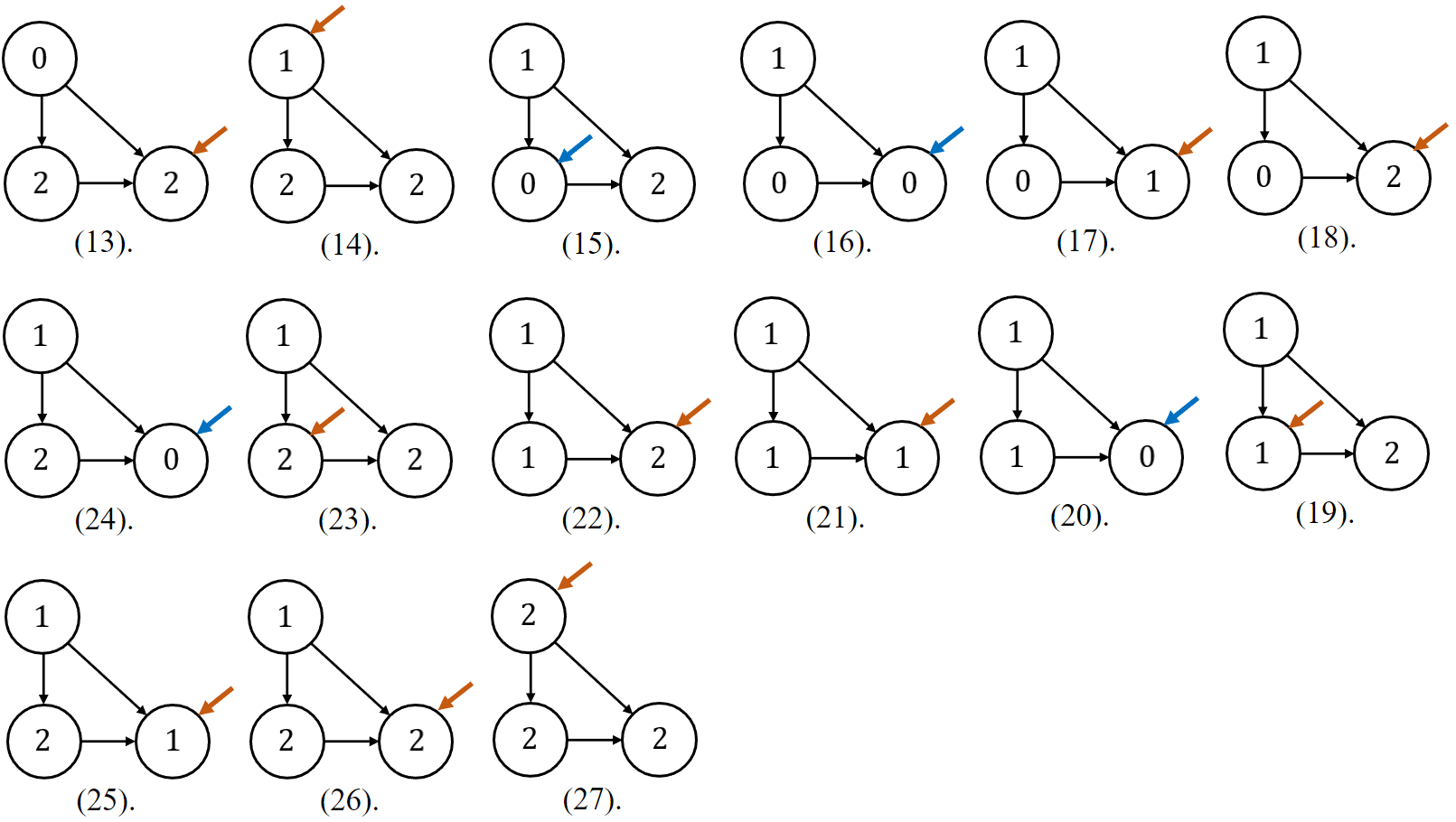}
\caption{(0). The example setup. (1).-(27). The execution procedure. The number in the circle indicates the gradient step $k_i$ of each node $\bm{y}_i$. The bold blue/red arrow indicates that the current node is under initialization/gradient ascent.}
\label{fig:eg}
\end{figure*}

\textbf{Theorem 3.4.}\textit{
After the procedure \textup{grad-dag($\bm{x},\bm{y}_0^{k_0},...,\bm{y}_i^{k_i}$)} in Alg.~\ref{alg:advancedag} executes, we have the return value $d\mathcal{L}(\bm{y}_0^{k_0},...,\bm{y}_i^{k_i},\bm{y}_{>i}^K)/d \bm{y}_i^{k_i}=\overleftarrow{\bm{y}_i}$.
}
\begin{proof}
Consider computing the target gradient with DAG $\mathcal{G}$. The $\bm{y}_i$'s gradient is composed of its own contribution to $\mathcal{L}$ in addition to the gradient from its children $\bm{y}_j\in\mathcal{C}(\bm{y}_i)$. Further, as we are considering the optimized children $\bm{y}_j^K$, we expand each child node $\bm{y}_j$ as Fig.~\ref{fig:cg2}. Then, we have:
\begin{align}
    \frac{d\mathcal{L}(\bm{y}_0^{k_0},...,\bm{y}_i^{k_i},\bm{y}_{>i}^K)}{d \bm{y}_i^{k_i}}&=\underbrace{\frac{\partial \mathcal{L}(\bm{y}_0^{k_0},...,\bm{y}_i^{k_i},\bm{y}_{>i}^K)}{\partial \bm{y}_i^{k_i}}}_{\textrm{known}} \notag\\+ \sum_{\bm{y}_j\in\mathcal{C}(\bm{y}_i)}(\sum_{k_j=0}^{K}&\underbrace{\frac{\partial \bm{y}_j^{k_j}}{\partial\bm{y}_i^{k_i}}}_{\textrm{Eq.~}\ref{eq:yw_gd}}\underbrace{\frac{d\mathcal{L}(\bm{y}_0^{k_0},...,\bm{y}_{j-1}^{k_{j-1}},\bm{y}_{\ge j}^K)}{d\bm{y}_j^{k_j}}}_{\textrm{Eq.~}\ref{eq:yKyk_gd}}).\label{eq:graddag}
\end{align}
The first term on the right-hand side of Eq.~\ref{eq:graddag} can be trivially evaluated. The $\partial\bm{a}_j^{k_j}/\partial\bm{a}_i^{k_i}$ can be evaluated as Eq.~\ref{eq:yw_gd} from the proof of Theorem.~\ref{th:2l}. The $ d\mathcal{L}(\bm{a}_0^{k_0},...,\bm{a}_{j-1}^{k_{j-1}},\bm{a}_{\ge j}^K)/d\bm{a}_j^{k_j}$ can also be iteratively expanded as Eq.~\ref{eq:yKyk_gd} from the proof of Theorem.~\ref{th:2l}. Then, the rest of proof follows Theorem.~\ref{th:2l} to expand Eq.~\ref{eq:yw_gd} and Eq.~\ref{eq:yKyk_gd}.

We highlight several key differences between Alg.~\ref{alg:advancedag} and Alg.~\ref{alg:advance2}:
\begin{itemize}
    \item The gradient evaluation of current node $\bm{y}_i$ requires gradient of its plural direct children $\bm{y}_j\in\mathcal{C}(\bm{y}_i)$, instead of the single child in 2-level case. The children traversal part of Eq.~\ref{eq:yKyk_gd} corresponds to the two extra for in Alg.~\ref{alg:advancedag}. 
    \item The gradient ascent update of child latent parameter $\bm{y}_j^{k_j+1}\leftarrow \bm{y}_j^{k_j}+\alpha d\mathcal{L}(\bm{y}_0^{k_0},...,\bm{y}_j^{k_j},\bm{y}_{>j}^K)/d\bm{y}_{j}^{k_j}$ can be conducted trivially only if $\mathcal{C}(\bm{y}_j)$ is empty, otherwise the gradient has to be evaluated recursively using Eq.~\ref{eq:graddag}. And this part corresponds to the recursive call in line 11 of Alg.~\ref{alg:advancedag}.
\end{itemize}
And the other part of Alg.~\ref{alg:advancedag} is the same as Alg.~\ref{alg:advance2}. So the rest of the proof follows Theorem.~\ref{th:2l}. Similarly, the Hessian-vector product can be approximated as Eq.~\ref{eq:hes}. However, this does not save Alg.~\ref{alg:advancedag} from an overall complexity of $\Theta(K^N)$.

To better understand Alg.~\ref{alg:advancedag}, we provide an example of the full procedure of its execution in Fig.~\ref{fig:eg}. The setup is as Fig.~\ref{fig:eg}.(0): we have $N=3$ latent $\bm{y}_1,\bm{y}_2,\bm{y}_3$ and gradient ascent step $K=2$, connected by a DAG shown in the figure.
\end{proof}
\section{Implementation Details}
\label{app:impl}
\subsection{Implementation Details of Density Estimation}
For density estimation experiment, we follow the training details of \citet{burda2015importance}. We adopt Adam \citep{Kingma2014AdamAM} optimizer with $\beta_1=0.9,\beta_2=0.999,\epsilon=1e-4$ and batch-size $20$. We adopt the same learning rate scheduler as \citep{burda2015importance} and train 3280 epochs for total. All likelihood results in Tab.~\ref{tab:savi} are evidence lowerbound. And the binarization of MNIST dataset follows \citep{Salakhutdinov2008OnTQ}.

\subsection{Implementation Details of Bit Allocation}
\begin{figure*}[htb]
\centering
    \includegraphics[width=\textwidth]{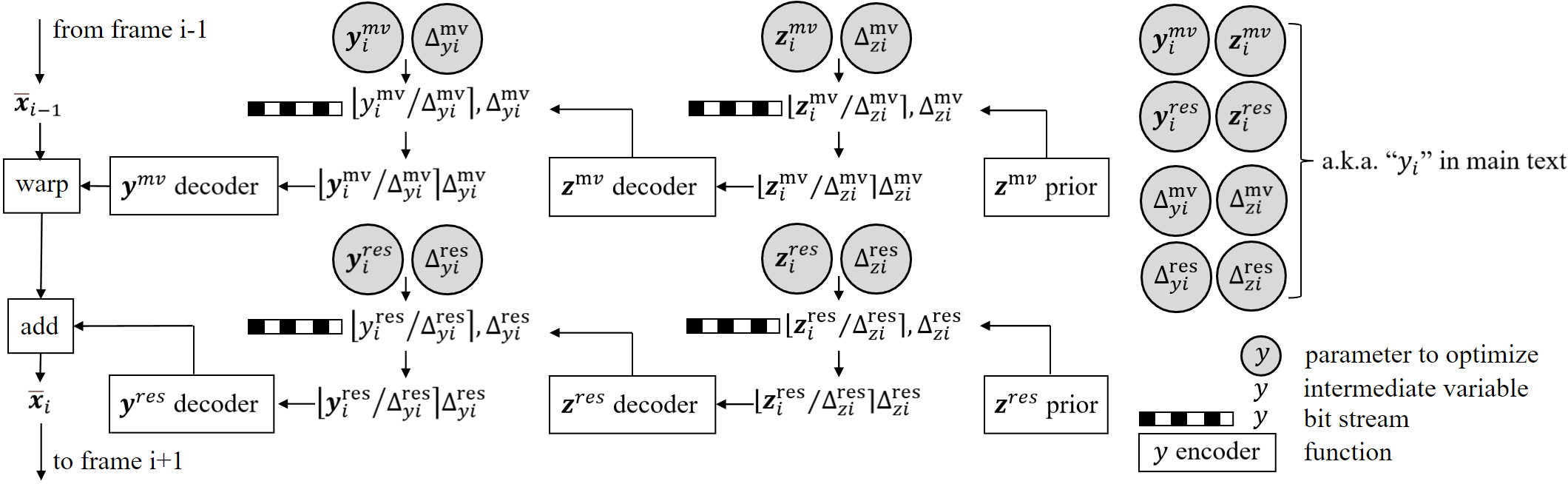}
    \caption{Overall framework of detailed implementation based on DVC \citep{lu2019dvc}. The DCVC's \citep{li2021deep} framework is very similar.}
    \label{fig:fraemwork}
\end{figure*}
In the main text and analysis, we use $\bm{y}_i$ to abstractly represent all the latent of frame $i$. While the practical implementation is more complicated. In fact, the actual latent is divided into 8 parts: the first level of residual latent $\bm{y}_i^{res}$, the second level of residual latent $\bm{z}_i^{res}$, the first level of optical flow latent $\bm{y}_i^{mv}$, and the second level of optical flow latent $\bm{z}_i^{mv}$. In addition to those 4 latent, we also include quantization stepsize $\Delta_{iy}^{res},\Delta_{iz}^{res},\Delta_{iy}^{mv},\Delta_{iz}^{mv}$ as \citet{Choi19} for DVC \citep{lu2019dvc} and DCVC \citep{li2021deep}. For HSTEM \citep{10.1145/3503161.3547845}, those quantization parameters are predicted from the latent $\bm{z}_i$, so we have not separately optimize them. This is shown in the overall framework diagram as Fig.~\ref{fig:fraemwork}. To re-parameterize the quantization, we adopt Stochastic Gumbel Annealing with the hyper-parameters as \citet{yang2020improving}.

\section{More Experimental Results}
\label{app:exp}
\subsection{R-D Performance}
\begin{figure*}[thb]
\centering
    \begin{minipage}[t]{0.40\textwidth}
    \includegraphics[width=\textwidth]{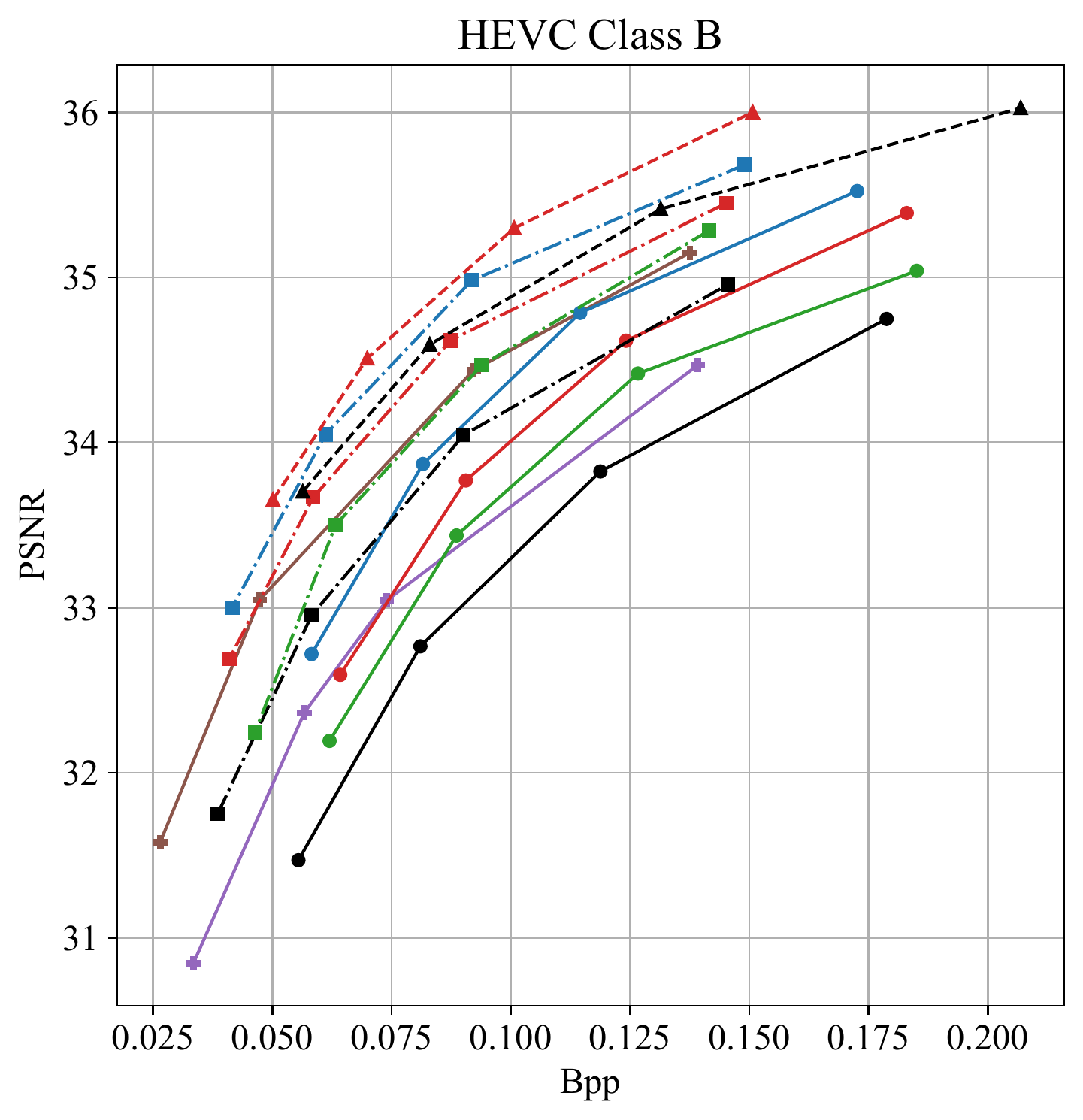}
    \end{minipage}%
    \begin{minipage}[t]{0.40\textwidth}
    \includegraphics[width=\textwidth]{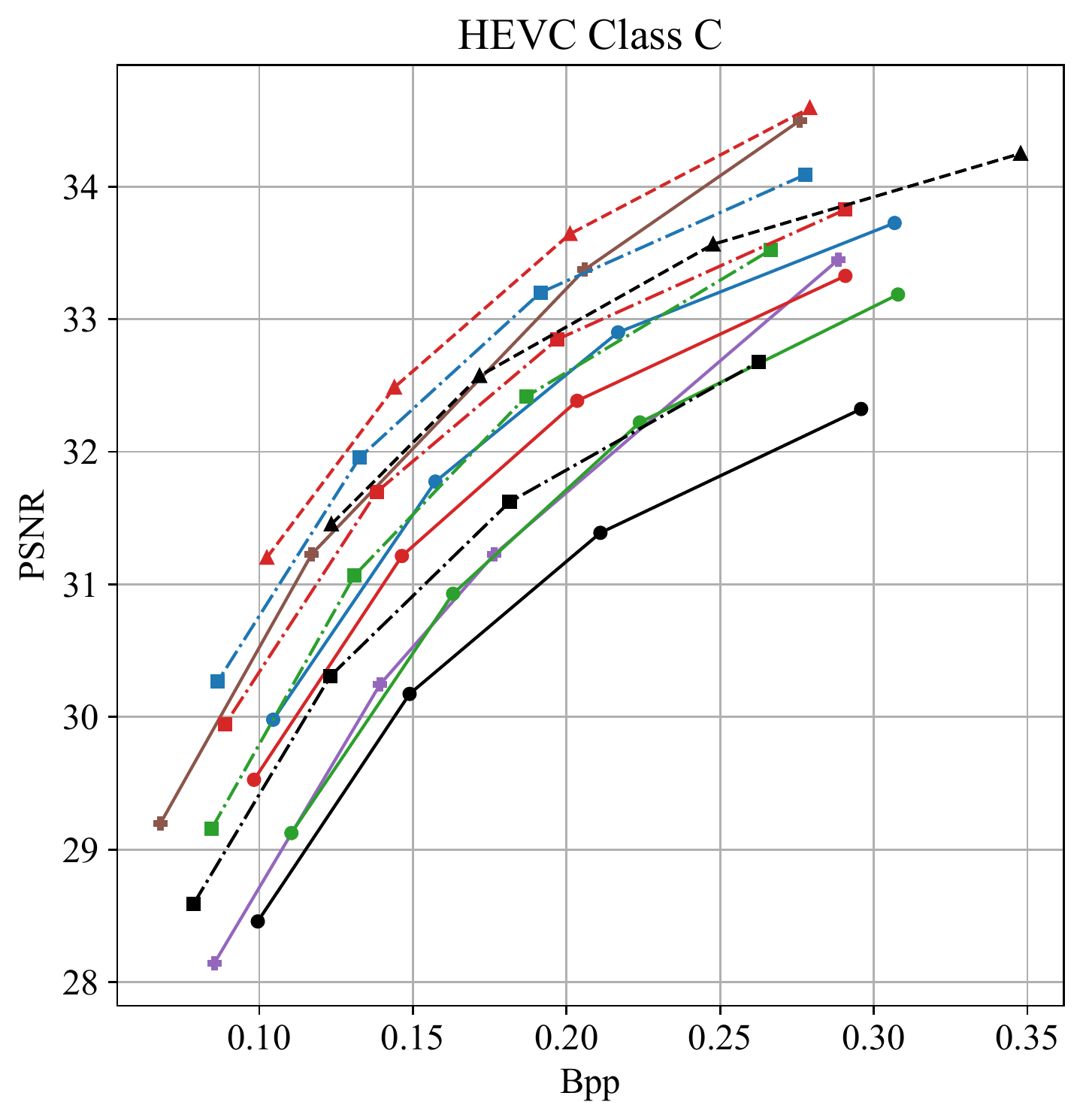}
    \end{minipage}%
    \vfill
    \begin{minipage}[t]{0.40\textwidth}
    \includegraphics[width=\textwidth]{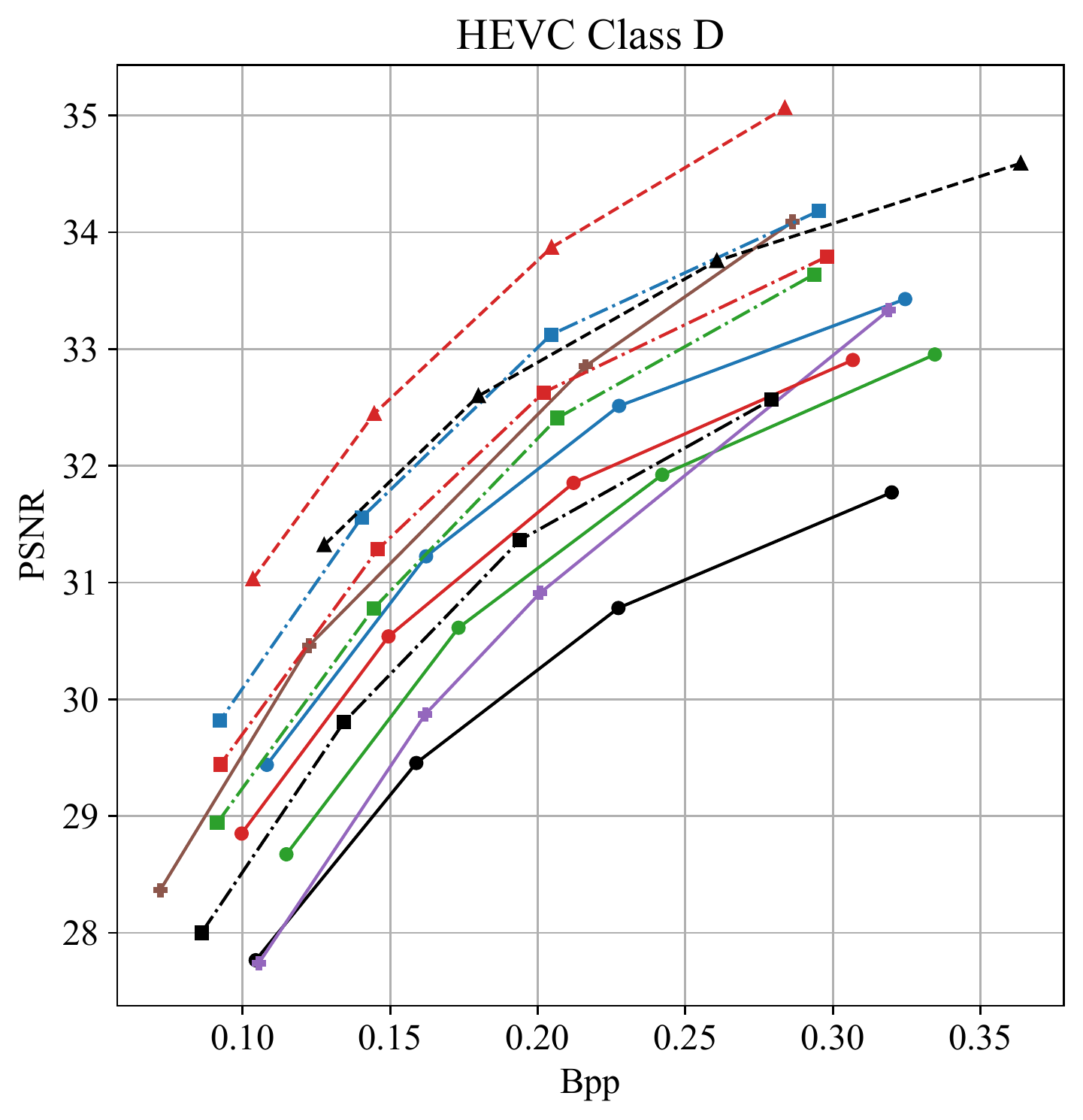}
    \end{minipage}%
    \begin{minipage}[t]{0.40\textwidth}
    \includegraphics[width=\textwidth]{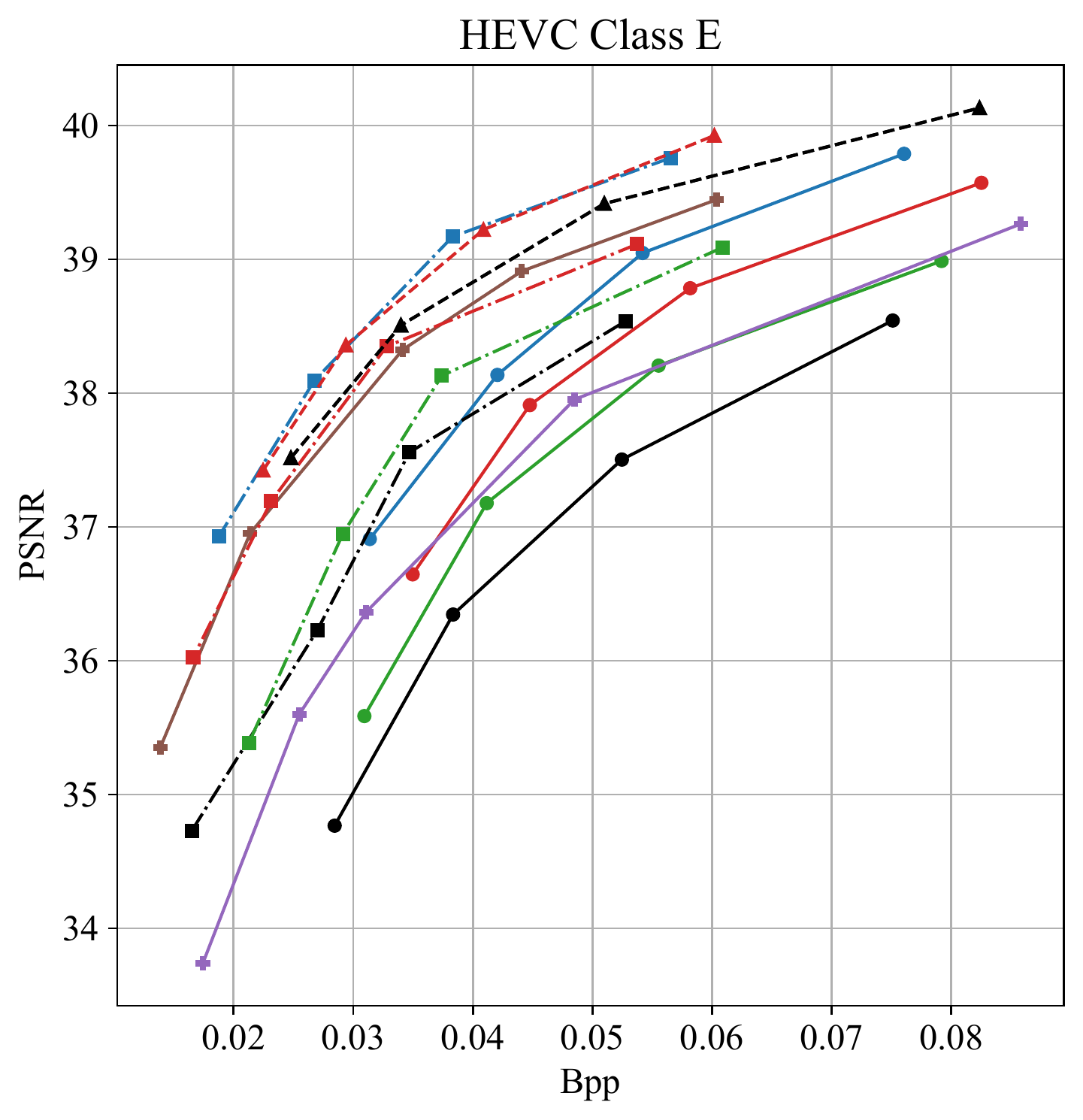}
    \end{minipage}%
    \vfill
    \begin{minipage}[t]{0.40\textwidth}
    \includegraphics[width=\textwidth]{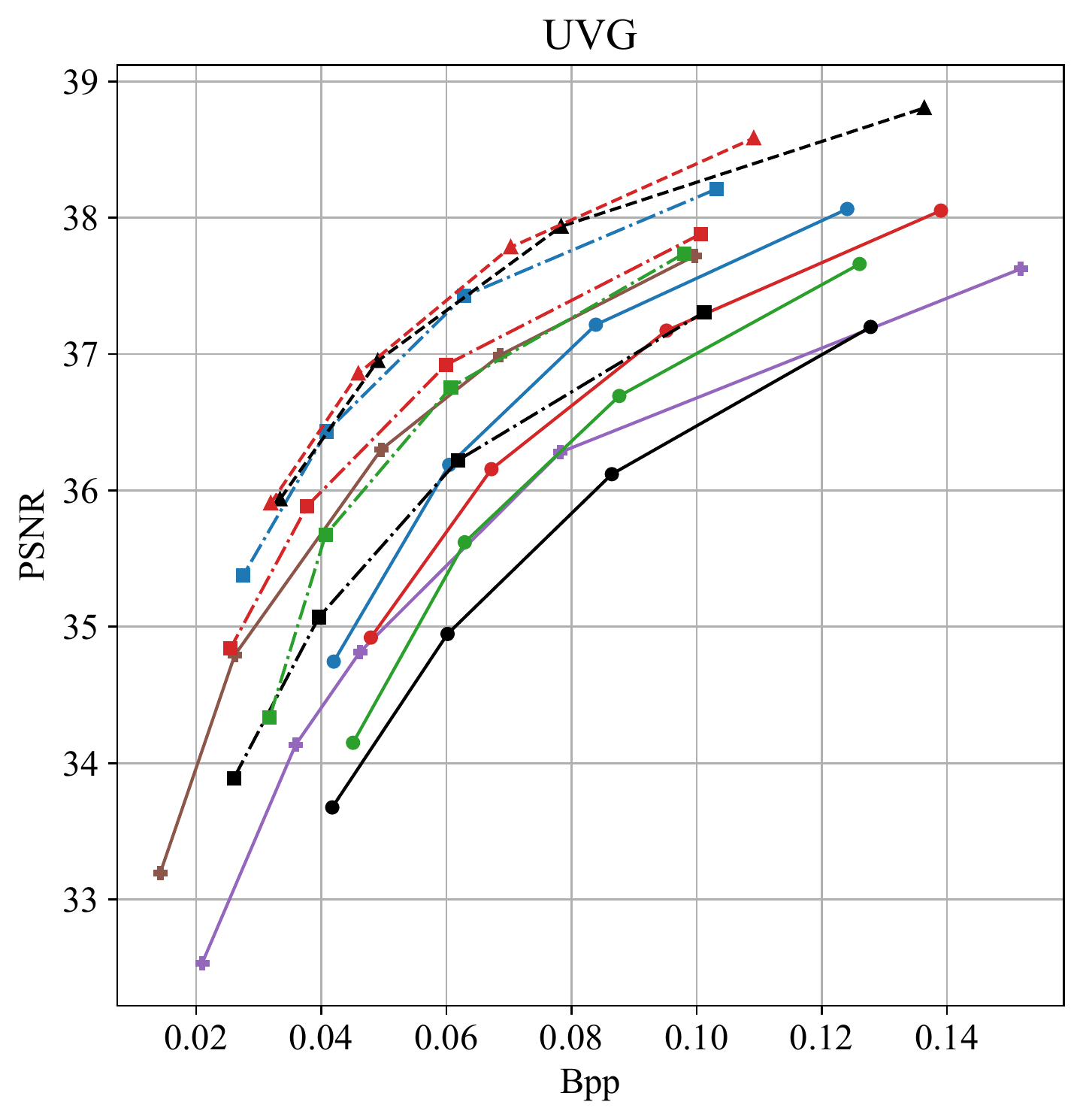}
    \end{minipage}%
    \begin{minipage}[t]{0.40\textwidth}    \includegraphics[width=\textwidth]{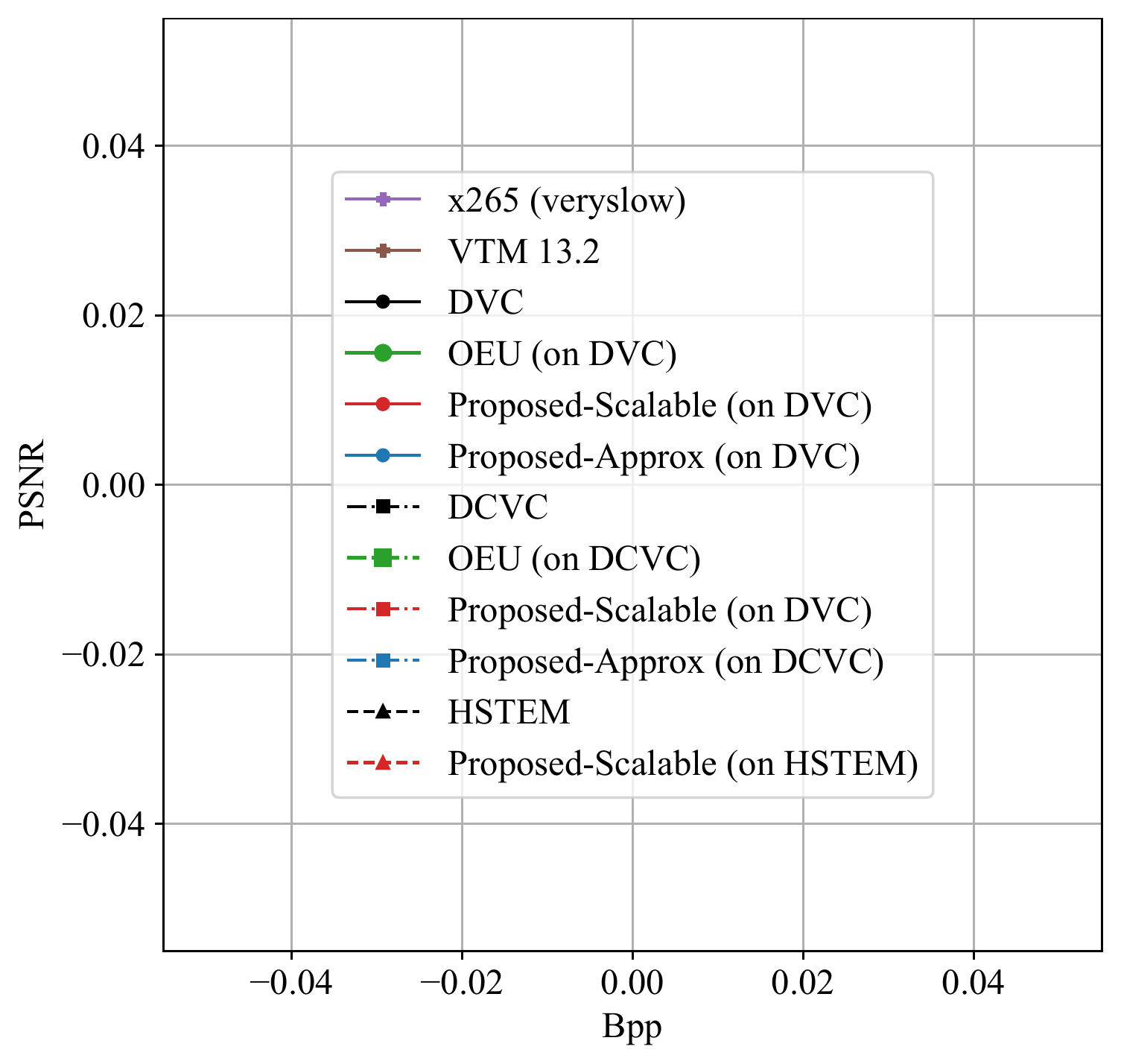}
    \end{minipage}%
    \caption{The R-D performance of our approach compared with baselines (w/o bit allocation) and other bit allocation approach.}
    \label{fig:rd_all}
\end{figure*}

Fig.~\ref{fig:rd_all} shows the R-D curves of proposed approach, other bit allocation approach \cite{lu2020content} and baselines \cite{lu2019dvc, li2021deep, 10.1145/3503161.3547845}, from which we observe that proposed approach has stable improvement upon three baselines and five datasets (HEVC B/C/D/E, UVG).

In addition to baselines and bit allocation results, we also show the R-D performance of traditional codecs. For H.265, we test x256 encoder with \textit{veryslow} preset. For H.266, we test VTM 13.2 with \textit{lowdelay P} preset. The command lines for x265 and VTM are as follows:
\begin{verbatim}
ffmpeg -y -pix_fmt yuv420p -s HxW
-r 50 -i src_yuv -vframes frame_cnt 
-c:v libx265 -preset veryslow
-x265-params "qp=qp:
keyint=gop_size" output_bin
\end{verbatim}
\begin{verbatim}
EncoderApp -c encoder_lowdelay_P_vtm.cfg
--QP=qp --InputFile=src_yuv
--BitstreamFile=output_bin 
--DecodingRefreshType=2
--InputBitDepth=8
--OutputBitDepth=8
--OutputBitDepthC=8
--InputChromaFormat=420
--Level=6.2 --FrameRate=50
--FramesToBeEncoded=frame_cnt
--SourceWidth=W
--SourceHeight=H
\end{verbatim}
And for x265, we test qp=$\{38,34,32,28,24\}$. For VTM 13.2, we test qp=$\{34,30,26,24,22\}$.

As Fig.~\ref{fig:rd_all} shows, with our bit allocation, the DCVC \citep{li2021deep} outperforms the latest traditional codec standard (VTM 13.2). The HSTEM baseline \citep{10.1145/3503161.3547845} already outperforms VTM 13.2, and our bit allocation makes this advantage more obvious.

\subsection{Qualitative Results}
\begin{figure*}[thb]
\centering
\includegraphics[width=0.8\linewidth]{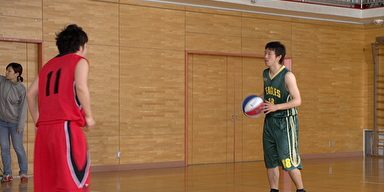}
\includegraphics[width=0.8\linewidth]{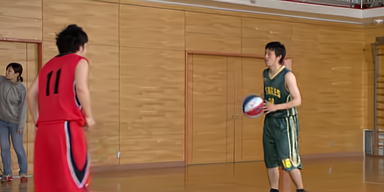}
\includegraphics[width=0.8\linewidth]{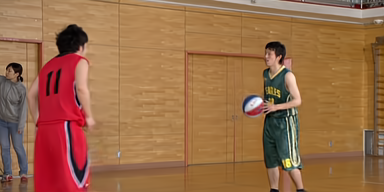}
\caption{Qualitative results using \textit{BasketballPass} of HEVC Class D. \textit{Top}. Original frame. \textit{Middle}. Baseline codec (DVC)'s reconstruction result with bpp $=0.148$ and PSNR $=33.06$dB. \textit{Bottom}. Proposed method's reconstruction result with bpp $=0.103$ and PSNR $=34.91$dB.}
\label{fig:qual1}
\end{figure*}

\begin{figure*}[thb]
\centering
\includegraphics[width=0.8\linewidth]{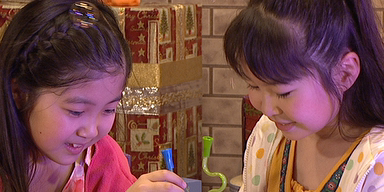}
\includegraphics[width=0.8\linewidth]{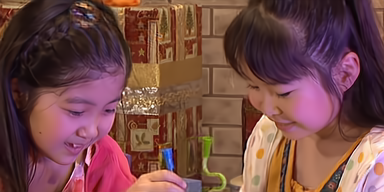}
\includegraphics[width=0.8\linewidth]{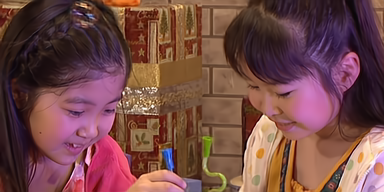}
\caption{Qualitative results using \textit{BlowingBubbles} of HEVC Class D. \textit{Top}. Original frame. \textit{Middle}. Baseline codec (DVC)'s reconstruction result with bpp $=0.206$ and PSNR $=30.71$dB. \textit{Bottom}. Proposed method's reconstruction result with bpp $=0.129$ and PSNR $=32.34$dB.}
\label{fig:qual2}
\end{figure*}

\begin{figure*}[thb]
\centering
\includegraphics[width=0.8\linewidth]{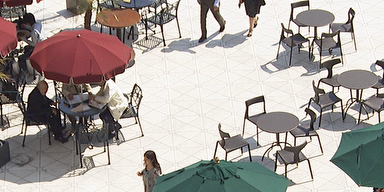}
\includegraphics[width=0.8\linewidth]{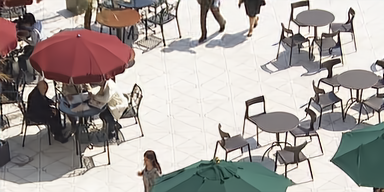}
\includegraphics[width=0.8\linewidth]{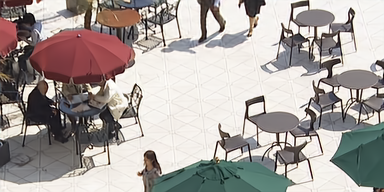}
\caption{Qualitative results using \textit{BQSquare} of HEVC Class D. \textit{Top}. Original frame. \textit{Middle}. Baseline codec (DVC)'s reconstruction result with bpp $=0.232$ and PSNR $=28.72$dB. \textit{Bottom}. Proposed method's reconstruction result with bpp $=0.128$ and PSNR $=30.87$dB.}
\label{fig:qual3}
\end{figure*}

\begin{figure*}[thb]
\centering
\includegraphics[width=0.8\linewidth]{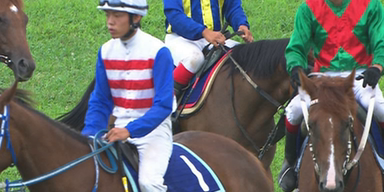}
\includegraphics[width=0.8\linewidth]{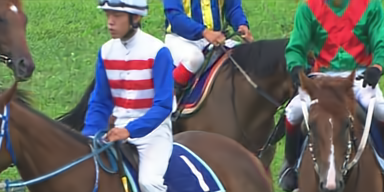}
\includegraphics[width=0.8\linewidth]{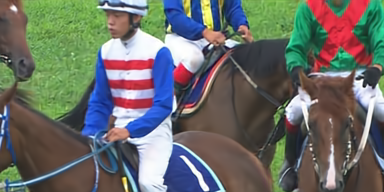}
\caption{Qualitative results using \textit{RaceHorses} of HEVC Class D. \textit{Top}. Original frame. \textit{Middle}. Baseline codec (DVC)'s reconstruction result with bpp $=0.448$ and PSNR $=30.48$dB. \textit{Bottom}. Proposed method's reconstruction result with bpp $=0.379$ and PSNR $=31.92$dB.}
\label{fig:qual4}
\end{figure*}

In Fig.~\ref{fig:qual1}, Fig.~\ref{fig:qual2}, Fig.~\ref{fig:qual3} and Fig.~\ref{fig:qual4}, we present the qualitative result of our approach compared with the baseline approach on HEVC Class D. We note that compared with the reconstruction frame of baseline approach, the reconstruction frame of our proposed approach preserves significantly more details with lower bitrate, and looks much more similar to the original frame. 

\section{More on Online Encoder Update (OEU)}
\label{app:oeu}
\subsection{Why OEU is a Frame-level Bit Allocation}
The online encoder update (OEU) of \citet{lu2020content} is proposed to resolve the error propagation. It fits into the line of works on encoder over-fitting \citep{cremer2018inference,van2021overfitting,zou20202} and does not fit into the SAVI framework. However, it can be seen as an intermediate point between fully-amortized variational inference (FAVI) and SAVI: FAVI uses the same inference model parameter $\phi$ for all frames; SAVI performs stochastic variational inference on variational posterior parameter $\bm{y}_i$ per frame, which is the finest-granularity finetune possible for latent variable model; And \citet{lu2020content} finetune inference model parameter $\phi_i$ per-frame. It is less computational expensive but also performs worse compared with our method based on SAVI (See Tab.~\ref{tab:lucmp}). However, it is more computational expensive and performs better compared with FAVI (or NVC without bit allocation).

\begin{table*}[htb]
\centering
\begin{tabular}{@{}lllll@{}}
\toprule
 & DVC & OEU & Proposed-Scalable (Sec.~\ref{sec:scalable}) &  Proposed-Approx (Sec.~\ref{sec:approx})\\ \midrule
Parameter & - & $\phi_i$ & $\bm{y}_i$ & $\bm{y}_i$ \\
Granularity & - & frame-level & pixel-level & pixel-level\\
\# of parameter & 0 & $\Theta(N|\phi|)$ & $\Theta(NHW)$ & $\Theta(NHW)$ \\
$K$ required & 0 & $\approx 50$ & $\approx 2000$ & $\approx 2000$ \\
R-D performance & poor & middle & good & very good \\
Temporal complexity & $\Theta(N)$ & $\Theta(KN)$ & $\Theta(KN)$ & $\Theta(KN)$ \\
Spatial complexity & $\Theta(1)$ & $\Theta(N)$ & $\Theta(1)$ & $\Theta(N)$ \\
Encoding time & fast & middle & slow & very slow\\
Decoding time & same & same & same & same\\
\bottomrule
\end{tabular}
\caption{A comparison between DVC (FAVI, w/o bit allocation), OEU \citep{lu2020content} and our proposed bit allocation (SAVI) approach. $N$ is the GoP size, $|\phi|$ is the number of encoder parameters, $H\times W$ is the number of pixels, $K$ is number of iteration.
}
\label{tab:lucmp}
\end{table*}

Although OEU does not fit into SAVI framework, still it can be written as Eq.~\ref{eq:a2:oeu1}, which is very similar to our SAVI-based implicit bit allocation with $\Delta_i$ removed and optimization target changed from $\bm{y}_i,\Delta_i$ to encoder parameter $\phi_i$. 
\begin{align}
    \phi_{1:K}^* \leftarrow \arg \max_{\phi_{1:K}} \mathcal{L} \label{eq:a2:oeu1}
\end{align}
This similarity also implies that \citet{lu2020content} is optimized towards the overall bit allocation target. Thus, it is also an optimal bit allocation. However, the encoder parameter $\phi_i$ is only sensitive to the location of a frame inside a GoP. It can not sense the pixel location and pixel-level reference structure. So the bit allocation of \citet{lu2020content} is limited to frame-level instead of pixel-level like ours.

Despite it is not pixel-level optimal bit allocation, OEU \citep{lu2020content} is indeed the true pioneer of bit allocation in NVC. 
However, the authors of \citet{lu2020content} have not mentioned the concept of bit allocation in the original paper, which makes it a secret pioneer of bit allocation for NVC until now.

\subsection{Why OEU is not an Error Propagation Aware Strategy}

One problem of NVC is that the frame quality drops rapidly with the frame index $t$ inside the GoP. Many works call this phenomena error propagation \citep{lu2020content,sun2021spatiotemporal,sheng2021temporal}, including the OEU. And the error propagation aware technique is subtly related to bit allocation.

\begin{figure}[H]
\centering
    \includegraphics[width=0.6\linewidth]{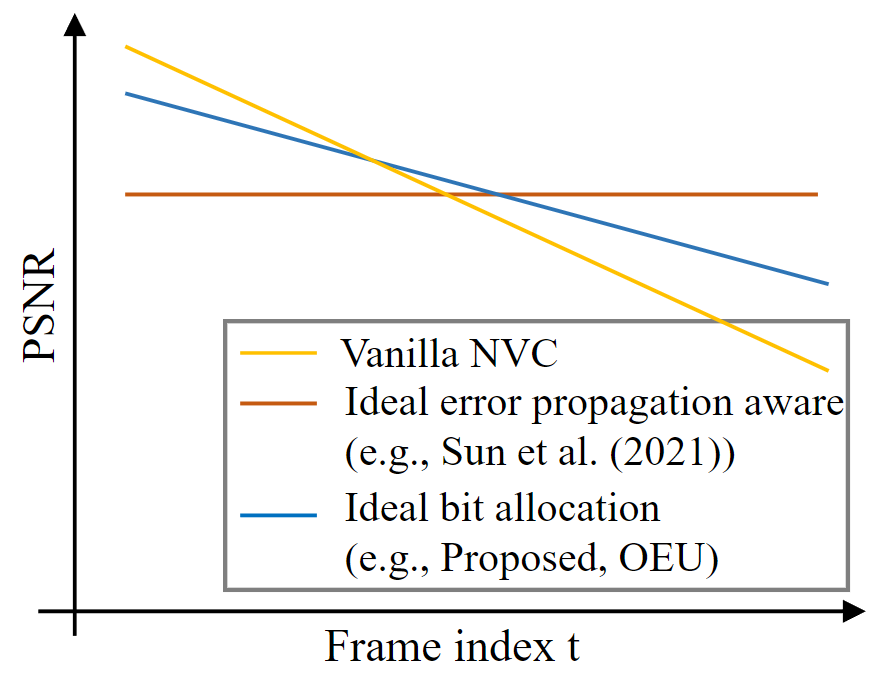}
    \caption{The comparison between ideal error propagation aware and ideal bit allocation.}
    \label{fig:epa}
\end{figure}

In this paper, we would like to distinguish error propagation aware technique and bit allocation formally: error propagation aware technique aims to solve the problem of quality drop with frame index $t$, and an ideal solution to error propagation should produce a horizontal quality-t curve, just like the red line in Fig.~\ref{fig:epa}. And \citet{sun2021spatiotemporal} aim to solve error propagation instead of bit allocation, as their frame quality is consistent to frame index $t$. However, the aim of bit allocation is to produce the best average R-D cost. And due to the frame reference structure, the frame quality of ideal bit allocation should drop with frame index $t$. As a result, the quality degrading rate of ideal bit allocation is less steep than vanilla NVC, but it almost never falls horizontal. 

From the old school bitrate control's perspective \citep{tagliasacchi2008minimum}, the target of avoiding error propagation is \textit{minVar}, which means that we want to minimize the quality fluctuation between frames. On the other hand, the target of bit allocation is \textit{minAvg}, which means that we want to minimize the average distortion under constrain of bitrate. And usually, those two targets are in odd with each other.

However, the results of a crippled error propagation aware method and an optimal bit allocation are similar: the degradation speed of quality-$t$ curve is decreased but not flattened. And thus, sometimes the result of an optimal bit allocation is likely to be recognised as an unsuccessful error propagation aware method. This similarity has disorientated \citet{lu2020content} to falsely classify themselves into error propagation aware strategy instead of bit allocation.

As we empirically show in Fig.~\ref{fig:rdram}.\textit{lower-left}, the quality-$t$ curves of both our proposed method and OEU are not horizontal. Instead, they are just less steep than the original DVC \citep{lu2019dvc}. This phenomena further verifies that OEU is essentially a bit allocation, instead of an error propagation aware strategy.
\section{More Discussion}
\label{sec:disc}
\subsection{More Limitations}
As we have stated in main text, the major limitation of our method is encoding complexity. Although the decoding time is not influenced, our method requires iterative gradient descent during encoding. Practically, our approach is limited to scenarios when R-D performance matters more than encoding time (e.g. content delivery network). Or it can also be used as a benchmark for research on faster bit allocation methods. 
\subsection{Ethics Statement}
Improving the R-D performance of NVC has valuable social impact. First, better R-D performance means less resources required for video storage and transmission. Considering the amount of video produced everyday, it is beneficial to the environment if we could save those resources by improving codec. Second, the traditional codec requires dedicated hardware decoder for efficient decoding, while NVC could utilize the universal neural accelerator. The improvement of neural codec can be readily deployed without hardware up-gradation.


\end{document}